\documentclass[lettersize,journal]{IEEEtran}
\usepackage{amsmath,amsfonts}
\usepackage{algorithmic}
\usepackage{array}
\usepackage[caption=false,font=normalsize,labelfont=sf,textfont=sf]{subfig}
\usepackage{textcomp}
\usepackage{stfloats}
\usepackage{url}
\usepackage{verbatim}
\usepackage{graphicx}
\usepackage{xcolor}
\usepackage{amssymb}
\usepackage{booktabs}   
\usepackage{multirow}   
\usepackage{booktabs} 
\usepackage{algorithm}
\usepackage{caption}
\usepackage{subfig}
\usepackage{amsmath}
\usepackage{amssymb}
\usepackage{amsthm}
\usepackage{pifont}

\def\BibTeX{{\rm B\kern-.05em{\sc i\kern-.025em b}\kern-.08em
		T\kern-.1667em\lower.7ex\hbox{E}\kern-.125emX}}
\usepackage{balance}
\begin{document}
	\title{Disentangled Double Machine Learning for Accurate Causal Effect Estimation}
	\author{Guodu Xiang, 
		       Kui Yu*, \textit{Senior Member, IEEE}, 
		       Yujie Wang, 
		       Richang Hong, \textit{Senior Member, IEEE},\\
		       Fuyuan Cao, 
		       and Jiye Liang, \textit{Fellow, IEEE}

		\thanks{
			Guodu Xiang, Kui Yu, Yujie Wang and Richang Hong are with the School of Computer Science and Information Engineering, Hefei University of Technology, Hefei {\rm 230601}, China. (e-mail: \{xgd600600, yujiewang\}@mail.hfut.edu.cn; ykui713@gmail.com;  hongrc@hfut.edu.cn) (*corresponding author: Kui Yu).}
			
			\thanks{Fuyuan Cao and Jiye Liang are with the School of Computer and Information Technology, Shanxi University, Taiyuan 030006, China. (e-mail: \{cfy, ljy\}@sxu.edu.cn).}
		}

	\markboth{Journal of \LaTeX\ Class Files,~Vol.~18, No.~9, September~2020}%
	{Disentangled Double Machine Learning}
	
	\maketitle
	
\begin{abstract}
Confounding bias is a key challenge in causal effect estimation from observational data. Double Machine Learning (DML) addresses this issue by estimating treatment and outcome nuisance functions, constructing treatment and outcome residuals, and estimating causal effects from the residuals. However, DML often produces biased and unstable estimates in high-dimensional or finite-sample scenarios. One reason is that DML estimates nuisance functions using all covariates without disentangling distinct latent factors, resulting in unreliable nuisance function estimation. Another is that imprecise nuisance estimation further introduces residual dependence between the treatment residual and the remaining outcome error, undermining the accuracy of causal effect estimates. To address these issues, in this paper, we propose \underline{D}isentangled \underline{D}ouble \underline{M}achine \underline{L}earning (DDML), a novel algorithm that integrates two key strategies. First, a causal role disentanglement strategy decomposes covariates into confounders, treatment-specific factors, and outcome-specific factors for enabling reliable nuisance function estimation. And second, a residual dependence orthogonalization strategy mitigates residual dependence caused by nuisance estimation errors for enhancing the precision of causal effect estimates. Experimental results on synthetic, semi-synthetic, and real-world datasets demonstrate that DDML significantly outperforms 13 state-of-the-art baseline algorithms in both MAE and RMSE.
\end{abstract}
	

\section{Introduction}
Causal effect estimation aims to calculate the impact of a treatment  variable $T$ on the outcome variable $Y$, and has been widely used in many fields \cite{feuerriegel2024causal, 10858762, 11344809}. For example, in labor economics, researchers may wish to evaluate whether participating in a job training program ($T$) affects an individual's income ($Y$). However, estimating causal effects from observational data remains challenging because of confounding bias, caused by confounders that simultaneously influence both the treatment and the outcome \cite{cheng2024data}.

To reduce confounding bias, traditional methods such as matching and propensity score-based approaches \cite{ridgeway2015propensity, wu2024matching} adjust observed covariates ($X$) between treatment and control groups. Nevertheless, these methods often struggle with high-dimensional covariates and with complex relationships. To address this limitation, representation learning-based methods \cite{shalit2017estimating, yao2018representation, bica2020estimating} have been proposed to learn balanced representations of covariates in latent spaces. Although these methods have achieved some empirical success, many existing approaches are primarily designed for binary treatment settings, which limits their applicability to more general treatment scenarios. To address this issue, Double Machine Learning (DML) \cite{chernozhukov2018double} provides a flexible framework for causal effect estimation. Specifically, DML first estimates two nuisance functions: the conditional expectation functions of the treatment and outcome, denoted by $\hat{m}(X)$ and $\hat{g}(X)$, where $m(X)=\mathbb{E}[T\mid X]$ and $g(X)=\mathbb{E}[Y\mid X]$. It then constructs the residuals of the treatment and outcome as $\tilde{T}=T-\hat{m}(X)$ and $\tilde{Y}=Y-\hat{g}(X)$, to remove the influence explained by $X$. The causal effect $\theta$ is estimated by regressing $\tilde{Y}$ on $\tilde{T}$. Built on Neyman orthogonality, DML has desirable theoretical debiasing properties for estimating causal effects.
\begin{figure}[ht]
	\centering	
	\subfloat[\rmfamily]{\includegraphics[width=0.475\linewidth]{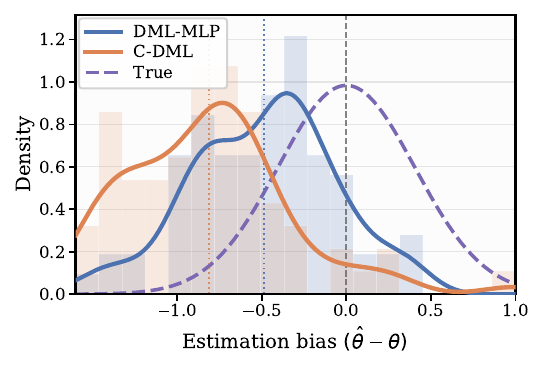}}
	\subfloat[\rmfamily]{\includegraphics[width=0.485\linewidth]{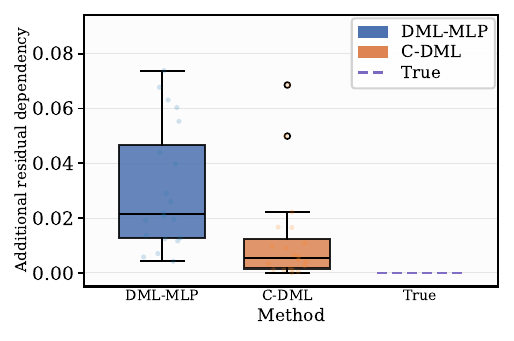}}
	\captionsetup{singlelinecheck=false, justification=justified}
	\caption{(a) Bias distributions of causal effect estimates obtained by DML-MLP (i.e., DML with a multi-layer perceptron estimator) and C-DML on the \textit{IHDP} dataset, where $\hat\theta$ and $\theta$ denote the estimated and true causal effects. (b) Additional dependency between the treatment residual and the remaining outcome error after removing the treatment effect under DML-MLP and C-DML.}
	\label{fig:motivation}
\end{figure}

However, this desirable orthogonal debiasing property is often difficult to be satisfied in practice, especially when a dataset has limited samples, high-dimensional covariates, or complex dependency relationships. As a result, DML-based methods may remain biased in real-world settings. As shown in Fig. \ref{fig:motivation} (a), we validate this issue on the semi-synthetic \textit{IHDP} dataset, where the bias distributions of the classic DML \cite{chernozhukov2018double} and its improved variant C-DML \cite{fingerhut2022coordinated} are shifted away from zero rather than centered around it, indicating that their estimation results are not consistent with the true causal effect. We find that this inaccuracy mainly arises from two reasons.

First, during nuisance function estimation, DML learns $m(X)$ and
$g(X)$ from the full covariates $X$, without separating variables with different causal roles. For example, when studying whether participating in a job training program ($T$) affects income ($Y$), education level and prior work experience may act as confounders that influence $T$ and $Y$, enrollment preference as a treatment-specific factor that influences only $T$, and job skill level as an outcome-specific factor that influences only $Y$. Therefore, treatment-specific factors may be introduced into the outcome model $g(X)$, while outcome-specific factors may be introduced into the treatment model $m(X)$, leading to nuisance estimation errors.

Second, during causal effect estimation, DML estimates the causal effect by regressing $\tilde{Y}$ on $\tilde{T}$, which relies on accurate nuisance estimation to remove covariate effects. Ideally, the remaining error term $\epsilon=\tilde{Y}-\theta \tilde{T}$ should be uncorrelated with $\tilde{T}$, i.e., $\mathbb{E}[\tilde{T} \epsilon ]=0$. However, when the nuisance functions are estimated inaccurately, some covariate effects may remain in the residuals. As a result, additional statistical dependence may remain between the treatment residual and the remaining outcome error after removing the treatment effect, which can be reflected by a violation of the moment condition $\mathbb{E}[\tilde{T}(\tilde{Y}-\theta \tilde{T})] \neq 0$, reducing the accuracy of causal effect estimation. As shown in Fig.~\ref{fig:motivation}~(b), our experiments on the \textit{IHDP} dataset provide empirical evidence for this extra dependence in both DML and C-DML. This indicates that the orthogonal moment condition may not be fully satisfied in practice.

To address the above issues, we propose DDML, a disentangled double machine learning algorithm that enhances the accuracy of causal effect estimation. Our main contributions are summarized as follows:
\begin{itemize}
	\item[$\boldsymbol{\bullet}$] We develop a Causal Role Disentanglement (CRD) strategy to decompose covariates into distinct causal factors, enabling more accurate estimation of nuisance functions under complex covariate dependency relationships.
	
	\item[$\boldsymbol{\bullet}$] We design a Residual Dependence Orthogonalization (RDO) constraint to mitigate residual dependence between the treatment residual and the remaining outcome error, improving the precision of causal effect estimation.
	
	\item[$\boldsymbol{\bullet}$] We conduct extensive experiments on synthetic, semi-synthetic, and real-world datasets to validate the effectiveness of DDML. Results show that DDML outperforms 13 state-of-the-art baselines, achieving average improvements of 57.45\% in MAE and 58.85\% in RMSE over the best competitor on synthetic datasets, and average improvements of 32.2\% in MAE and 29.3\% in RMSE on semi-synthetic and real-world datasets.
\end{itemize}

\section{Related Work}\label{R_W}
We focus on the potential outcomes model (POM) \cite{shalit2017estimating, yao2021survey}-based causal effect estimation methods which can be broadly categorized into traditional methods, representation learning methods, and debiased estimation methods.

Traditional methods mitigate confounding bias primarily through covariate adjustment and propensity score modeling, with representative approaches including matching \cite{li2016matching, wu2024matching}, inverse probability weighting (IPW) \cite{wang2024debiased, ma2020robust}, doubly robust estimation (DR) \cite{funk2011doubly, zhang2022doubly}, and tree-based models \cite{chipman2006bayesian, chipman2010bart}. However, in high-dimensional settings with complex nonlinearities, these methods suffer from the curse of dimensionality.

To overcome these limitations, representation learning methods map covariates into a latent space and balance these learned representations across different treatment groups. Classic balancing-based methods, such as BNN \cite{johansson2016learning} and CFRNet \cite{shalit2017estimating}, estimate causal effects by reducing distributional discrepancies between treatment and control groups in the learned representation space. Related extensions further improve representation balancing through local similarity preservation and contrastive learning \cite{yao2018representation, wang2025proximity}. Representative generative and adversarial methods, such as CEVAE \cite{louizos2017causal} and GANITE \cite{yoon2018ganite}, improve causal effect estimation by recovering latent confounders or predicting counterfactual outcomes. Subsequent variants \cite{bica2020estimating, du2021adversarial, qiang2025glcn} of these models have further enhanced robustness against complex observational biases. In addition, methods such as D$^{2}$VD \cite{kuang2017treatment}, DR-CFR \cite{hassanpour2019learning}, and related extensions \cite{wu2023learning, huang2024dignet, zhangcounterfactual}, improve causal effect estimation by incorporating additional constraints into representation learning, enhancing the model's robustness. However, existing representation learning methods are mainly designed for binary treatments and rely on representation balancing, limiting their applicability to more general treatment settings.

Double Machine Learning (DML) \cite{chernozhukov2018double} is a flexible framework for causal effect estimation with strong debiasing properties. It estimates causal effects by residualizing the treatment and outcome variables with their corresponding nuisance functions. Combined with Neyman orthogonality and cross-fitting, DML enables more reliable causal effect estimation. Building on this framework, several extensions have been proposed. DML-ID \cite{jung2021estimating} extends DML to general causal diagrams. C-DML \cite{fingerhut2022coordinated} improves finite-sample estimation through coordinated learning. Auto-DML \cite{chernozhukov2022automatic} further automates bias correction and avoids manual derivation of orthogonal score functions. However, these methods do not distinguish covariate roles in nuisance estimation, leading to inaccurate effect estimates in practice.

\section{Preliminaries}\label{P_D}
\subsection{Notations and Assumptions}
Let the observed dataset be $\mathcal{D}=\{(X_i,T_i,Y_i)\}_{i=1}^N$, where $i$ indexes individuals and $N$ is the sample size. For each individual $i$, $X_i\in\mathcal{X}\subset\mathbb{R}^d$ denotes a $d$-dimensional vector of observed covariates, $T_i$ denotes the treatment, and $Y_i\in\mathbb{R}$ denotes the observed outcome. In this paper, we consider both binary and continuous treatment settings. For simplicity, we take the binary treatment setting as an example, where $T_i\in\{0,1\}$, with $T_i=1$ indicating treatment and $T_i=0$ indicating control. Under the POM, let $Y_i(1)$ and $Y_i(0)$ denote the potential outcomes of individual $i$ under treatment and control, respectively, such that the observed outcome satisfies $Y_i = T_iY_i(1) + (1-T_i)Y_i(0)$. The individual treatment effect (ITE) is defined as $\tau_i = Y_i(1)-Y_i(0)$, but it is not directly estimated because one potential outcome is always missing. Our target estimand is the average treatment effect (ATE), $\theta = \mathbb{E}[Y_i(1)-Y_i(0)]$, where the expectation is taken over the population. The continuous treatment setting considered later is formulated under the partially linear model, where the target parameter corresponds to the ATE of treatment on the outcome. The main notations are summarized in Table \ref{tab:notations}.
\begin{table}[h]
	\centering
	\caption{Notations and Their Meanings}
	\label{tab:notations}
	\renewcommand{\arraystretch}{1}
	\begin{tabular}{l l}
		\toprule
		\textbf{Notation} & \textbf{Meaning} \\
		\midrule
		$\mathcal{P}(Y, T, X)$ & Data distribution\\
		$i$ & Individual \\
		$N$ & Sample size \\
		$X\in\mathbb{R}^d$ & Covariates (Features) \\
		$d$ & Feature dimension \\
		$T\in\mathbb{R}$ & Treatment \\
		$Y\in\mathbb{R}$ & Outcome \\
		$Y(0),Y(1)$ & Potential outcomes of control and treatment \\
		$\mathcal{D}$ & Dataset $\{(X_i,T_i,Y_i)\}_{i=1}^N$ \\
		$\tau_i$ & Individual causal effect, ITE \\
		$\theta$ & Average treatment effect, ATE \\
		$g(\cdot)$ & Outcome nuisance\\
		$m(\cdot)$ & Treatment nuisance\\
		$\eta=\{g,m\}$ & Nuisance functions \\
		$\psi(\cdot)$ & Orthogonal score \\
		$\mathcal{I}_k$ & The $k$-th fold \\
		$\mathcal{I}_{\mathrm{tra}}$ & Training fold $\mathcal{D}\setminus\mathcal{I}_k$ \\
		$\mathcal{I}_{\mathrm{hold}}$ & Held-out fold, $\mathcal{I}_k$ \\
		$\Phi$ & Disentangled encoder \\
		$\Psi$ & Prediction network \\
		$(Z_c,Z_t,Z_y)$ & Latent representations for different causal roles \\
		$\hat{T},\hat{Y}$ & Cross-fitted nuisance predictions \\
		$K$ & The number of folds \\
		$\lambda_{\mathrm{dis}}$ & Weight of $\mathcal{L}_{\mathrm{dis}}$ \\
		$\lambda_{\mathrm{ort}}$ & Weight of $\mathcal{L}_{\mathrm{ort}}$ \\
		$\Theta$ & Regularization term\\
		\bottomrule
	\end{tabular}
\end{table}

The ATE can be estimated from the data distribution $\mathcal{P}(X, T, Y)$ under the following assumptions.

\noindent \textbf{Assumption 3.1.} Consistency:
\textit{For each individual $i$, the observed outcome equals the potential outcome under the received treatment, i.e., $Y_i = Y_i(T_i)$ }(\textit{equivalently, if $T_i=t$ then $Y_i=Y_i(t)$}).

\noindent \textbf{Assumption 3.2.} Stable Unit Treatment Value Assumption (SUTVA):
\textit{There is no interference between units and no hidden versions of treatment, so each unit's potential outcomes depend only on its own treatment assignment.}

\noindent \textbf{Assumption 3.3.} Unconfoundedness: \textit{Treatment assignment is independent of the potential outcomes conditional on observed covariates, i.e., $T\perp\!\!\!\perp\left\{ Y(t) \right\} \mid X$, implying that all confounding variables are observed.}

\noindent \textbf{Assumption 3.4.} Positivity: \textit{For every individual, the probability of receiving either treatment is positive, i.e., $0<P(T=t \mid X)<1$, ensuring that all relevant subgroups are represented in treatment groups.}
\begin{figure*}[!htbp]
	\centering
	\includegraphics[width=0.78\linewidth]{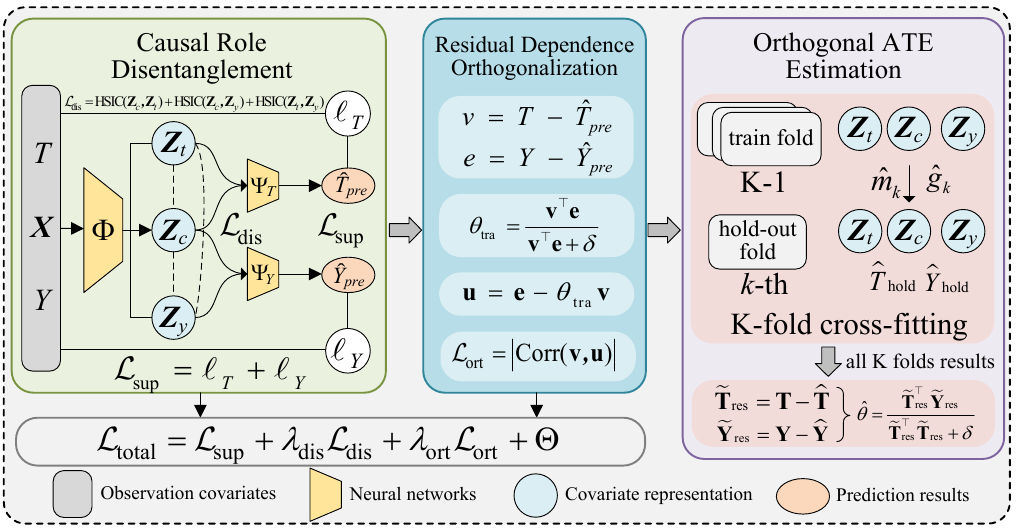}
	\caption{Overview of the DDML algorithm.}
	\label{fig:DDML}
\end{figure*}
\subsection{Double Machine Learning}
Under Assumptions 3.1--3.4, the target causal effect $\theta_0$ can be estimated from the observed data distribution $\mathcal{P}(X,T,Y)$. We consider the partially linear model (PLM), where $N$ independent and identically distributed samples $(X_i,T_i,Y_i)$ are drawn from the same distribution:
\begin{equation}
	\label{eq:plm}
	\begin{aligned}
		Y &= \theta_0 T + f_0(X) + \zeta, \quad \mathbb{E}[\zeta \mid X,T] = 0,\\
		T &= m_0(X) + \varepsilon, \quad \mathbb{E}[\varepsilon \mid X] = 0,
	\end{aligned}
\end{equation}
where $\theta_0$ is the target causal effect, $m_0(X)=\mathbb{E}[T\mid X]$, and $g_0(X)=\mathbb{E}[Y\mid X]$. Under the PLM, we have
\begin{equation}
	g_0(X)=\theta_0 m_0(X)+f_0(X).
\end{equation}

DML estimates $\theta_0$ through an orthogonal score function $\psi(W;\theta,\eta)$, where $W=(X,T,Y)$ and $\eta=\{g,m\}$ denotes the nuisance functions. At the population level, define the residualized treatment and outcome under the true nuisance functions as
\begin{equation}
	\tilde T = T-m_0(X), \qquad \tilde Y = Y-g_0(X).
\end{equation}
Then the PLM implies
\begin{equation}
	\tilde Y = \theta_0 \tilde T + \zeta,
\end{equation}
which yields the ideal population moment condition
\begin{equation}
	\mathbb E\big[(\tilde Y-\theta_0 \tilde T)\tilde T\big]=0.
\end{equation}
Motivated by this oracle moment equation, the Neyman-orthogonal score is given by
\begin{equation}
	\label{eq:orth_score}
	\psi(W;\theta,\eta)=\big(Y-g(X)-\theta(T-m(X))\big)\big(T-m(X)\big),
\end{equation}
where $g(X)$ and $m(X)$ are generic nuisance functions. This score satisfies
the Neyman orthogonality condition at the true parameter values
$(\theta_0,\eta_0)$, with $\eta_0=\{g_0,m_0\}$, namely,
\begin{equation}
	\left.\frac{\partial}{\partial \eta}\mathbb{E}\big[\psi(W;\theta_0,\eta)\big]\right|_{\eta=\eta_0}=0.
\end{equation}

To mitigate overfitting and enable valid inference when estimating nuisance functions with flexible machine learning models, DML employs $K$-fold cross-fitting. Specifically, the sample indices $\{1,\ldots,N\}$ are randomly partitioned into $K$ mutually exclusive folds $\{\mathcal{I}_k\}_{k=1}^K$. For each fold $k$, the nuisance functions $\hat g_k$ and $\hat m_k$ are estimated on the complementary sample $\{1,\ldots,N\}\setminus \mathcal{I}_k$ and evaluated on the held-out fold $\mathcal{I}_k$. For each $i\in\mathcal{I}_k$, the out-of-fold residuals are constructed as:
\begin{equation}
	\tilde T_i = T_i-\hat m_k(X_i), \qquad
	\tilde Y_i = Y_i-\hat g_k(X_i).
\end{equation}
Aggregating the out-of-fold residuals across all folds yields the DML estimator:
\begin{equation}
	\hat{\theta}
	=
	\frac{\tilde{\mathbf T}^{\top}\tilde{\mathbf Y}}{\tilde{\mathbf T}^{\top}\tilde{\mathbf T}}.
\end{equation}
Therefore, DML estimates the causal effect by regressing the cross-fitted residualized outcome on the cross-fitted residualized treatment under the orthogonal moment condition.

\section{The Proposed DDML Method}\label{DD_ML}

To address the practical bias of double machine learning caused by mixed covariate roles and residual dependence, we propose a Disentangled Double Machine Learning (DDML) algorithm. As illustrated in Fig.~\ref{fig:DDML}, DDML consists of three procedures. First, to resolve the mixed causal roles of covariates in nuisance learning, we design a Causal Role Disentanglement (CRD) strategy to separate confounders, treatment-specific factors, and outcome-specific factors for improving the accuracy of nuisance estimation. Second, to mitigate the remaining dependence between the treatment residual and the remaining outcome error, we develop a Residual Dependence Orthogonalization (RDO) constraint. Finally, we integrate these two components into a cross-fitted procedure for orthogonal ATE estimation. The pseudocode of DDML is provided in Algorithm \ref{alg:ddml}.
\begin{algorithm}[htbp]
	\caption{DDML: Disentangled Double Machine Learning}
	\label{alg:ddml}
	\begin{algorithmic}[1]
		\STATE \textbf{Input:} dataset $\mathcal{D}=\{(X_i,T_i,Y_i)\}_{i=1}^N$; folds $K$; hyperparameters $\lambda_{\mathrm{dis}},\lambda_{\mathrm{ort}}$; constant $\delta$
		\STATE \textbf{Output:} $\hat{\theta}$
		\STATE Randomly split indices $\{1,\dots,N\}$ into $K$ folds $\{\mathcal{I}_k\}_{k=1}^K$
		\STATE $\hat{\mathbf{T}}\leftarrow\emptyset$; $\hat{\mathbf{Y}}\leftarrow\emptyset$
		\FOR{$k=1$ \TO $K$}
		\STATE $\mathcal{I}_{\mathrm{hold}}\leftarrow \mathcal{I}_k$, \quad $\mathcal{I}_{\mathrm{tra}}\leftarrow \{1,\dots,N\}\setminus\mathcal{I}_k$
		\STATE Initialize encoder $\Phi$ and prediction networks $\Psi$
		\STATE \textbf{\# Causal Role Disentanglement}
		\STATE $(\mathbf{Z}_c,\mathbf{Z}_t,\mathbf{Z}_y)= \Phi(\mathbf{X}[\mathcal{I}_{\mathrm{tra}}])$
		\STATE $\mathcal{L}_{\mathrm{dis}}= \mathrm{HSIC}(\mathbf{Z}_c,\mathbf{Z}_t)+\mathrm{HSIC}(\mathbf{Z}_c,\mathbf{Z}_y)+\mathrm{HSIC}(\mathbf{Z}_t,\mathbf{Z}_y)$
		\STATE $\hat{\mathbf{T}}_{\mathrm{pre}}= \Psi_T(\mathbf{Z}_c,\mathbf{Z}_t)$, $\hat{\mathbf{Y}}_{\mathrm{pre}}= \Psi_Y(\mathbf{Z}_c,\mathbf{Z}_y)$
		\STATE $\mathcal{L}_{\mathrm{sup}}= \ell_T(\hat{\mathbf{T}}_{\mathrm{pre}},\mathbf{T}[\mathcal{I}_{\mathrm{tra}}]) + \ell_Y(\hat{\mathbf{Y}}_{\mathrm{pre}},\mathbf{Y}[\mathcal{I}_{\mathrm{tra}}])$
		
		\STATE \textbf{\# Residual Dependence Orthogonalization}
		\STATE $\mathbf{v}= \mathbf{T}[\mathcal{I}_{\mathrm{tra}}]-\hat{\mathbf{T}}_{\mathrm{pre}}$
		\STATE $\mathbf{e}= \mathbf{Y}[\mathcal{I}_{\mathrm{tra}}]-\hat{\mathbf{Y}}_{\mathrm{pre}}$
		\STATE $\theta_{\mathrm{tra}}= \dfrac{\mathbf{v}^{\top}\mathbf{e}}{\mathbf{v}^{\top}\mathbf{v}+\delta}$
		\STATE $\mathbf{u}= \mathbf{e}-\theta_{\mathrm{tra}}\mathbf{v}$
		\STATE $\mathcal{L}_{\mathrm{ort}}= \left|\mathrm{Corr}(\mathbf{v},\mathbf{u})\right|$
		\STATE $\mathcal{L}_{\mathrm{total}}= \mathcal{L}_{\mathrm{sup}}+\lambda_{\mathrm{dis}}\mathcal{L}_{\mathrm{dis}}+\lambda_{\mathrm{ort}}\mathcal{L}_{\mathrm{ort}}+\Theta$
		\STATE Update $(\Phi,\Psi)$ by minimizing $\mathcal{L}_{\mathrm{total}}$
		\STATE \textbf{\# Orthogonal ATE Estimation}
		\STATE $(\mathbf{Z}_c^{\mathrm{tra}},\mathbf{Z}_t^{\mathrm{tra}},\mathbf{Z}_y^{\mathrm{tra}})= \Phi_k(\mathbf{X}[\mathcal{I}_{\mathrm{tra}}])$
		\STATE $(\mathbf{Z}_c^{\mathrm{hold}},\mathbf{Z}_t^{\mathrm{hold}},\mathbf{Z}_y^{\mathrm{hold}})= \Phi_k(\mathbf{X}[\mathcal{I}_{\mathrm{hold}}])$
		\STATE Fit treatment model $\hat{m}_k$ on $(\mathbf{Z}_c^{\mathrm{tra}},\mathbf{Z}_t^{\mathrm{tra}},\mathbf{T}[\mathcal{I}_{\mathrm{tra}}])$
		\STATE Fit outcome model $\hat{g}_k$ on $(\mathbf{Z}_c^{\mathrm{tra}},\mathbf{Z}_y^{\mathrm{tra}},\mathbf{Y}[\mathcal{I}_{\mathrm{tra}}])$
		\STATE $\hat{\mathbf{T}}[\mathcal{I}_{\mathrm{hold}}]= \hat{m}_k(\mathbf{Z}_c^{\mathrm{hold}},\mathbf{Z}_t^{\mathrm{hold}})$;
		\\ $\hat{\mathbf{Y}}[\mathcal{I}_{\mathrm{hold}}]= \hat{g}_k(\mathbf{Z}_c^{\mathrm{hold}},\mathbf{Z}_y^{\mathrm{hold}})$
		\STATE $\hat{\mathbf{T}}=\hat{\mathbf{T}}\cup \hat{\mathbf{T}}[\mathcal{I}_{\mathrm{hold}}]$ ; \quad$\hat{\mathbf{Y}}=\hat{\mathbf{Y}}\cup \hat{\mathbf{Y}}[\mathcal{I}_{\mathrm{hold}}]$ ;
		\ENDFOR
		\STATE $\mathbf{\tilde T}_{\mathrm{res}}= \mathbf{T}-\hat{\mathbf{T}}$, $\mathbf{\tilde Y}_{\mathrm{res}}= \mathbf{Y}-\hat{\mathbf{Y}}$
		\STATE $\hat{\theta}= \dfrac{\mathbf{\tilde T}_{\mathrm{res}}^{\top}\mathbf{\tilde Y}_{\mathrm{res}}}{\mathbf{\tilde T}_{\mathrm{res}}^{\top}\mathbf{\tilde T}_{\mathrm{res}}+\delta}$
	\end{algorithmic}
\end{algorithm}

\subsection{\textbf{Causal Role Disentanglement}}
\begin{figure}[h]
	\centering
	\includegraphics[width=0.86\linewidth]{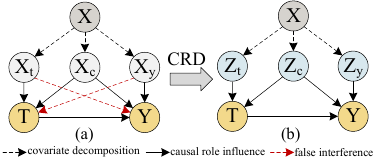}
	\caption{(a) The observed covariates $X$ contain treatment-specific variables $X_t$, confounders $X_c$, and outcome-specific variables $X_y$. (b) CRD disentangles $X$ into three latent representations $Z_t$, $Z_c$ and $Z_y$.}
	\label{fig:Causal_Structure}
\end{figure}

As shown in Fig.~\ref{fig:Causal_Structure} (a), the observed covariates $X$ may contain confounders $X_c$, treatment-specific factors $X_t$, and outcome-specific factors $X_y$. Directly learning the treatment and outcome nuisance functions from the original covariates may mix these causal roles, impairing nuisance estimation and reducing the accuracy of causal effect estimation. To address this issue, we propose a Causal Role Disentanglement (CRD) strategy, which disentangles $X$ into three role-specific latent representations for nuisance learning as in Fig.~\ref{fig:Causal_Structure} (b).

For the $k$-th fold of data, we use $\mathcal{I}_{\mathrm{hold}}=\mathcal{I}_k$ as the held-out sample and $\mathcal{I}_{\mathrm{tra}}=\{1,\ldots,N\}\setminus \mathcal{I}_k$ as the training sample. On $\mathcal{I}_{\mathrm{tra}}$, we first initialize a representation function $\Phi$, which maps the original covariates into three disentangled latent representations, as defined in Eq.~(\ref{eq:encoder}):
\begin{equation}
	(\mathbf{Z}_c,\mathbf{Z}_t,\mathbf{Z}_y)= \Phi(\mathbf{X}[\mathcal{I}_{\mathrm{tra}}]),
	\label{eq:encoder}
\end{equation}
where $\mathbf{Z}_c$ captures confounding information, $\mathbf{Z}_t$ captures treatment-specific information, and $\mathbf{Z}_y$ captures outcome-specific information.

To disentangle the mixing of information from different causal roles in the learned representations, we decompose the representation space into three subspaces. As shown in Eq. (\ref{L_HSIC}), we employ the \textit{Hilbert-Schmidt Independence Criterion} (HSIC) to penalize the statistical dependence among $(\mathbf{Z}_c,\mathbf{Z}_t,\mathbf{Z}_y)$, which promotes the disentanglement of representations associated with different causal roles.
\begin{equation}
	\mathcal{L}_{\mathrm{dis}}
	=
	\mathrm{HSIC}(\mathbf{Z}_c,\mathbf{Z}_t)
	+
	\mathrm{HSIC}(\mathbf{Z}_c,\mathbf{Z}_y)
	+
	\mathrm{HSIC}(\mathbf{Z}_t,\mathbf{Z}_y).
	\label{L_HSIC}
\end{equation}

Compared with linear correlation constraints, HSIC can capture higher-order nonlinear dependencies among latent representations, making it more suitable for disentangling complex causal factors and providing clearer and more structured representations for nuisance estimation.

To ensure that the learned representations remain informative for both treatment and outcome modeling, we further design a task-specific prediction network. Specifically, we employ a dual-head prediction network $\Psi$ to predict $T$ and $Y$ based on these representations, as shown in Eq. (\ref{eq:dual_head}):
\begin{equation}
	\hat{\mathbf{T}}_{\mathrm{pre}}= \Psi_T(\mathbf{Z}_c,\mathbf{Z}_t), \quad \hat{\mathbf{Y}}_{\mathrm{pre}}= \Psi_Y(\mathbf{Z}_c,\mathbf{Z}_y).
	\label{eq:dual_head}
\end{equation}
Here, $\hat{\mathbf{T}}_{\mathrm{pre}}$ denotes the predicted treatment, which is a probability for binary $T$, or a real-valued prediction for continuous $T$, and $\hat{\mathbf{Y}}_{\mathrm{pre}}$ denotes the predicted outcome. We train the prediction heads using task-specific supervised losses, as shown in Eq. (\ref{eq:lsup}). 
\begin{equation}
	\mathcal{L}_{\mathrm{sup}}
	=
	\ell_T\!\left(\hat{\mathbf{T}}_{\mathrm{pre}},\mathbf{T}[\mathcal{I}_{\mathrm{tra}}]\right)
	+
	\ell_Y\!\left(\hat{\mathbf{Y}}_{\mathrm{pre}},\mathbf{Y}[\mathcal{I}_{\mathrm{tra}}]\right).
	\label{eq:lsup}
\end{equation}

We set $\ell_Y(\cdot,\cdot)$ to mean squared error (MSE) for real-valued outcomes. For the treatment loss $\ell_T(\cdot,\cdot)$, we instantiate it as binary cross-entropy when $T\in\{0,1\}$, and as mean squared error (MSE) when $T$ is continuous. 

Since the dual-head prediction network is built upon different subspace representations, this task-specific design encourages the learned representations to encode treatment-related and outcome-related information into separate latent components. In this way, representation separation together with task-specific supervision promotes role-specific disentanglement without additional constraints. Moreover, the CRD module is applicable to both binary and continuous treatment settings.

\subsection{\textbf{Residual Dependence Orthogonalization}}
After obtaining disentangled representations through CRD, we further design a RDO constraint to reduce residual dependence between the treatment residual and the remaining outcome error. Specifically, as shown in Eq. (\ref{eq:pre_res}), we first calculate the treatment and outcome residuals on the training set:
\begin{equation}
	\mathbf{v}=\mathbf{T}[\mathcal{I}_{\mathrm{tra}}]-\hat{\mathbf{T}}_{\mathrm{pre}},\quad
	\mathbf{e}=\mathbf{Y}[\mathcal{I}_{\mathrm{tra}}]-\hat{\mathbf{Y}}_{\mathrm{pre}}.
	\label{eq:pre_res}
\end{equation}

We then compute the treatment effect $\theta_{\mathrm{tra}}$ on the training set  according to Eq.~(\ref{eq:theta_tra}):
\begin{equation}
	\theta_{\mathrm{tra}}
	=
	\frac{\mathbf{v}^{\top}\mathbf{e}}{\mathbf{v}^{\top}\mathbf{v}+\delta},
	\label{eq:theta_tra}
\end{equation}
where $\delta>0$ is a small constant for numerical stability. Based on $\theta_{\mathrm{tra}}$, we define the orthogonalized outcome residual in Eq. (\ref{eq:u_def}):
\begin{equation}
	\mathbf{u}=\mathbf{e}-\theta_{\mathrm{tra}}\mathbf{v}.
	\label{eq:u_def}
\end{equation}

To further mitigate the remaining residual dependence, we penalize the absolute Pearson correlation between $\mathbf{v}$ and $\mathbf{u}$:
\begin{equation}
	\mathcal{L}_{\mathrm{ort}}
	=
	\left|
	\mathrm{Corr}(\mathbf{v},\mathbf{u})
	\right|,
	\label{eq:lorth}
\end{equation}
where $\mathrm{Corr}(\mathbf{v},\mathbf{u})$ denotes the sample Pearson correlation, defined as
$\mathrm{Corr}(\mathbf{v},\mathbf{u})=
\frac{\sum_{i}(v_i-\bar v)(u_i-\bar u)}
{\sqrt{\sum_{i}(v_i-\bar v)^2}\sqrt{\sum_{i}(u_i-\bar u)^2}+\epsilon}$,
with a small $\epsilon>0$ for numerical stability.

Under this residual construction, $\mathbf{v}$ represents the treatment residual, while $\mathbf{u}$ denotes the orthogonalized outcome residual obtained by removing the linear contribution of the treatment residual from the outcome residual. Therefore, minimizing $|\mathrm{Corr}(\mathbf{v},\mathbf{u})|$ directly reduces residual dependence between the treatment residual and the orthogonalized outcome residual, improving residual orthogonality and enhancing the accuracy of subsequent orthogonal ATE estimation.

Building on the CRD strategy for causal role disentanglement and the RDO constraint for residual orthogonalization, we jointly optimize the representation encoder and task-specific prediction heads through the following total objective:
\begin{equation}
	\mathcal{L}_{\mathrm{total}}
	=
	\mathcal{L}_{\mathrm{sup}}
	+
	\lambda_{\mathrm{dis}}\mathcal{L}_{\mathrm{dis}}
	+
	\lambda_{\mathrm{ort}}\mathcal{L}_{\mathrm{ort}}
	+
	\Theta,
	\label{eq:ltotal}
\end{equation}
where $\lambda_{\mathrm{dis}}$ and $\lambda_{\mathrm{ort}}$ control the weights of disentanglement and orthogonality regularization, respectively, and $\Theta$ denotes standard regularization terms. In this objective, $\mathcal{L}_{\mathrm{sup}}$ preserves task-relevant predictive loss for nuisance estimation, $\mathcal{L}_{\mathrm{dis}}$ encourages the separation of latent factors with different causal roles, and $\mathcal{L}_{\mathrm{ort}}$ reduces the residual dependence that may remain after residualization. As a result, the learned representation is not only predictive but also structurally aligned with causal roles and better suited for orthogonal effect estimation. Sections~\ref{res:vs} and \ref{res:as} further demonstrate through visualization analysis and ablation studies that both CRD and RDO are effective, and that each module contributes to the performance of DDML.

\subsection{\textbf{Orthogonal ATE Estimation}}
Through the CRD and RDO procedures, we learn a disentangled and orthogonally regularized representation function. Based on this, we perform orthogonal ATE estimation by integrating causal role-guided nuisance function estimation with cross-fitting. Specifically, under the cross-fitting framework, for the $k$-th fold data, we learn a representation function $\Phi_{k}$ on $\mathcal{I}_{\mathrm{tra}}$. We apply $\Phi_{k}$ to the covariates in both $\mathcal{I}_{\mathrm{tra}}$ and  $\mathcal{I}_{\mathrm{hold}}$ to produce three disentangled latent representations, as shown in Eq. (\ref{Z_tra_hold}):
\begin{equation}
	\begin{aligned}
		(\mathbf{Z}_c^{\mathrm{tra}},\mathbf{Z}_t^{\mathrm{tra}},\mathbf{Z}_y^{\mathrm{tra}})
		&=
		\Phi_{k}(\mathbf{X}[\mathcal{I}_{\mathrm{tra}}]), \\
		(\mathbf{Z}_c^{\mathrm{hold}},\mathbf{Z}_t^{\mathrm{hold}},\mathbf{Z}_y^{\mathrm{hold}})
		&=
		\Phi_{k}(\mathbf{X}[\mathcal{I}_{\mathrm{hold}}]).
	\end{aligned}
	\label{Z_tra_hold}
\end{equation}

Then, we fit the nuisance functions $\hat{m}_k$ and $\hat{g}_k$ on the training fold $\mathcal{I}_{\mathrm{tra}}$ according to the disentangled causal roles in the learned representation, as shown in Eq. (\ref{tra:m_g}):
\begin{equation}
	\hat{m}_k:\,(\mathbf{Z}_c^{\mathrm{tra}},\mathbf{Z}_t^{\mathrm{tra}})\mapsto \mathbf{T}[\mathcal{I}_{\mathrm{tra}}],
	\quad
	\hat{g}_k:\,(\mathbf{Z}_c^{\mathrm{tra}},\mathbf{Z}_y^{\mathrm{tra}})\mapsto \mathbf{Y}[\mathcal{I}_{\mathrm{tra}}].
	\label{tra:m_g}
\end{equation}

This design encourages each nuisance function to be built on causally relevant representations, which reduces nuisance estimation errors.
As shown in Eq. (\ref{hold:m_g}), the nuisance models $\hat{m}_k$ and $\hat{g}_k$ fitted on the training fold are applied to the held-out fold to generate out-of-fold predictions:
\begin{equation}
	\hat{\mathbf{T}}[\mathcal{I}_{\mathrm{hold}}] = \hat{m}_k(\mathbf{Z}_c^{\mathrm{hold}},\mathbf{Z}_t^{\mathrm{hold}}),
	\hat{\mathbf{Y}}[\mathcal{I}_{\mathrm{hold}}] = \hat{g}_k(\mathbf{Z}_c^{\mathrm{hold}},\mathbf{Z}_y^{\mathrm{hold}}).
	\label{hold:m_g}
\end{equation}

After completing the $K$-fold cross-fitting procedure, we obtain nuisance predictions $(\hat{\mathbf{T}},\hat{\mathbf{Y}})$ for all samples. Based on these predictions, we define the treatment and outcome residuals as given in Eq.~(\ref{residuals}):
\begin{equation}
	\tilde{\mathbf T}_{\mathrm{res}}=\mathbf{T}-\hat{\mathbf{T}},
	\quad
	\tilde{\mathbf Y}_{\mathrm{res}}=\mathbf{Y}-\hat{\mathbf{Y}}.
	\label{residuals}
\end{equation}

As shown in Eq.~(\ref{eq:ddml_theta}), the estimated causal effect, denoted by $\hat{\theta}$, is obtained via residual-on-residual regression using the residuals of treatment and outcome from all samples:
\begin{equation}
	\hat{\theta}
	=
	\dfrac{\tilde{\mathbf T}_{\mathrm{res}}^{\top}\tilde{\mathbf Y}_{\mathrm{res}}}{\tilde{\mathbf T}_{\mathrm{res}}^{\top}\tilde{\mathbf T}_{\mathrm{res}}+\delta},
	\label{eq:ddml_theta}
\end{equation}
where $\delta>0$ is a small constant for numerical stability.

\section{Theoretical Analysis}\label{DDML_theory}
This section presents the theoretical analysis of DDML. We first show that performing nuisance estimation on the disentangled representation, rather than on the original covariates, preserves the identification of the target causal effect. We then prove that the score function induced by the disentangled representation continues to satisfy Neyman orthogonality. Finally, we establish the consistency of the DDML estimator under cross-fitting. To support the above theoretical analysis, we make the following assumptions.

\noindent \textbf{Assumption 5.1.} Role-preserving representation sufficiency. \textit{Let $Z=(Z_c,Z_t,Z_y)=\Phi(X)$ be the learned representation. Assume that:}
\begin{equation}
	\mathbb{E}[T\mid X]=\mathbb{E}[T\mid Z_c,Z_t],
	\quad
	\mathbb{E}[Y\mid X]=\mathbb{E}[Y\mid Z_c,Z_y].
	\label{eq:role_sufficiency}
\end{equation}
That is, the learned representation preserves the treatment-related and outcome-related information required for nuisance estimation in separate subspaces.

\noindent \textbf{Assumption 5.2.} Regularity and non-degeneracy.
\textit{Let}
$m(Z)=m(Z_c,Z_t)=\mathbb{E}[T\mid Z_c,Z_t]$
\textit{and}
$g(Z)=g(Z_c,Z_y)=\mathbb{E}[Y\mid Z_c,Z_y]$.
\textit{Assume that}
\begin{equation}
	\mathbb{E}\!\left[
	Y^2 + T^2 + (Y-g(Z))^2 + (T-m(Z))^2
	\right] < \infty,
	\label{eq:reg1}
\end{equation}
\begin{equation}
	\mathbb{E}\!\left[(T-m(Z))^2\right] \ge c > 0.
	\label{eq:reg2}
\end{equation}
These conditions ensure that the orthogonal moment is well-defined and that the residualized treatment remains non-degenerate.

\noindent \textbf{Assumption 5.3.} Cross-fitted nuisance consistency with dependence control.
\textit{Let $\hat m$ and $\hat g$ be cross-fitted estimators of the nuisance functions $m_0$ and $g_0$, respectively, where each prediction is obtained from a model trained on a sample that excludes the corresponding observation. Assume that}
\begin{equation}
	\|\hat m-m_0\|_{L_2}=o_p(1),
	\|\hat g-g_0\|_{L_2}=o_p(1),
	\label{eq:nuisance_consistency}
\end{equation}
\textit{and}
\begin{equation}
	\mathbb{E}\!\left[(\hat m-m_0)(\hat g-g_0)\right]=o_p(1).
	\label{eq:dependence_control}
\end{equation}
Here, $\|h\|_{L_2}:=(\mathbb{E}[h^2])^{1/2}$ denotes the $L_2$ norm, and $o_p(1)$ means convergence to zero in probability. That is, both nuisance estimators are simultaneously consistent in $L_2$ norm, and the interaction between their estimation errors becomes asymptotically negligible.

Under Assumptions 5.1 and 5.2, Proposition 5.1 shows that the causal effect remains identifiable in the disentangled representation space, and the ATE $\theta$ can be estimated through orthogonal moments on $Z$.

\noindent \textbf{Proposition 5.1.} \textit{Under Assumptions 5.1 and 5.2, let $m(Z)=m(Z_c,Z_t)$
	\textit{and}
	$g(Z)=g(Z_c,Z_y)$. Then the true causal effect $\theta_0$ satisfies:}
\begin{equation}
	\theta_0
	=
	\frac{
		\mathbb{E}\big[(T-m(Z))(Y-g(Z))\big]
	}{
		\mathbb{E}\big[(T-m(Z))^2\big]
	}.
	\label{eq:theta_representation}
\end{equation}
\begin{proof}
	\begingroup\small
	By the partially linear model,
	\begin{align*}
		Y=\theta_0 T + f_0(X)+\zeta,
		\mathbb{E}[\zeta\mid X,T]=0,
	\end{align*}
	where $f(X)$ denotes the structural regression term. Let
	\begin{align*}
		Z=\Phi(X),
		m(Z)=m(Z_c,Z_t),
		g(Z)=g(Z_c,Z_y).
	\end{align*}
	
	Under Assumption 5.1, the treatment and outcome nuisance functions depend on $X$
	through the corresponding representation subspaces. Hence,
	\begin{align*}
		\mathbb{E}[T\mid X]
		&=
		\mathbb{E}[T\mid Z_c,Z_t]
		=:m(Z),\\
		\mathbb{E}[Y\mid X]
		&=
		\mathbb{E}[Y\mid Z_c,Z_y]
		=:g(Z).
	\end{align*}
	
	By Assumption 5.1 and the tower property, we have
	\begin{align*}
		\mathbb{E}[T\mid Z]=m(Z),
		\mathbb{E}[Y\mid Z]=g(Z).
	\end{align*}
	
	Taking conditional expectation of the PLM given $X$ yields
	\begin{align*}
		\mathbb{E}[Y\mid X]
		=
		\theta_0\,\mathbb{E}[T\mid X]+f_0(X).
	\end{align*}
	
	Substituting $\mathbb{E}[Y\mid X]=g(Z)$ and $\mathbb{E}[T\mid X]=m(Z)$ gives
	\begin{align*}
		g(Z)=\theta_0 m(Z)+f_0(X),
	\end{align*}
	and therefore
	\begin{align*}
		f(X)=g(Z)-\theta_0 m(Z).
	\end{align*}
	
	Substituting this identity into the PLM gives
	\begin{align*}
		Y-g(Z)
		=
		\theta_0\bigl(T-m(Z)\bigr)+\zeta.
	\end{align*}
	
	Multiplying both sides by $T-m(Z)$ and taking expectations yields
	\begin{align*}
		\mathbb{E}\big[(T-m(Z))(Y-g(Z))\big]
		&=
		\theta_0\,\mathbb{E}\big[(T-m(Z))^2\big]\\
		&\quad+
		\mathbb{E}\big[(T-m(Z))\zeta\big].
	\end{align*}
	
	For the last term, since $T-m(Z)$ is measurable with respect to $(Z,T)$ and
	\begin{align*}
		\mathbb{E}[\zeta\mid Z,T]
		=
		\mathbb{E}\!\left[\mathbb{E}[\zeta\mid X,T]\mid Z,T\right]
		=0,
	\end{align*}
	we have
	\begin{align*}
		\mathbb{E}\big[(T-m(Z))\zeta\big]
		=
		\mathbb{E}\!\left[(T-m(Z))\,\mathbb{E}[\zeta\mid Z,T]\right]
		=
		0.
	\end{align*}
	
	Therefore,
	\begin{align*}
		\mathbb{E}\big[(T-m(Z))(Y-g(Z))\big]
		=
		\theta_0\,\mathbb{E}\big[(T-m(Z))^2\big].
	\end{align*}
	
	Assumption 5.2 ensures that $\mathbb{E}\big[(T-m(Z))^2\big]>0$, rearranging the above equality yields Eq. \eqref{eq:theta_representation}.
	\endgroup
\end{proof}

Proposition 5.1 provides a formal identification basis for CRD by showing that the causal effect remains recoverable from the disentangled representation. Building on this, Theorem 5.1 establishes that the score constructed on the learned representation remains Neyman-orthogonal.

\noindent \textbf{Theorem 5.1.} \textit{Under Assumptions 5.1 and 5.2, let $m(Z)=m(Z_c,Z_t)$
	\textit{and}
	$g(Z)=g(Z_c,Z_y)$, and consider the score}
\begin{align*}
	\psi_Z(W;\theta,\eta)
	=
	\big(Y-g(Z)-\theta(T-m(Z))\big)\big(T-m(Z)\big),
	\label{eq:scoreZ}
\end{align*}
\textit{where $\eta=(g,m)$. Then, at the true parameter $(\theta_0,\eta_0)$,}
\begin{equation}
	\left.
	\frac{\partial}{\partial \eta}\,
	\mathbb{E}\big[\psi_Z(W;\theta_0,\eta)\big]
	\right|_{\eta=\eta_0}
	=0.
	\label{eq:orthogonality}
\end{equation}
\begin{proof}
	Let $(\theta_0,\eta_0)$ denote the true parameter pair, where
	$
	\eta_0=(g_0,m_0),
	m_0(Z)=m_0(Z_c,Z_t),
	g_0(Z)=g_0(Z_c,Z_y).
	$
	Under Assumption 5.1, the treatment and outcome conditional means depend on $X$
	through the corresponding representation subspaces. Thus,
	\begin{align*}
		\mathbb{E}[T\mid X]
		&=
		\mathbb{E}[T\mid Z_c,Z_t]
		=:m_0(Z),\\
		\mathbb{E}[Y\mid X]
		&=
		\mathbb{E}[Y\mid Z_c,Z_y]
		=:g_0(Z).
	\end{align*}
	Since $Z=(\mathbf{Z}_c,\mathbf{Z}_t,\mathbf{Z}_y)$ is a function of $X$, iterated expectations imply
	\begin{align*}
		\mathbb{E}[T\mid Z]=m_0(Z),
		\mathbb{E}[Y\mid Z]=g_0(Z).
	\end{align*}
	
	Define the score based on the learned representation:
	\begin{equation*}
		\psi_Z(W;\theta,\eta)
		=
		\big(Y-g(Z)-\theta(T-m(Z))\big)\big(T-m(Z)\big).
	\end{equation*}
	
	Let
	\begin{equation*}
		v:=T-m_0(Z).
	\end{equation*}
	
	Then, by construction,
	\begin{equation*}
		\mathbb{E}[v\mid Z]
		=
		\mathbb{E}[T\mid Z]-m_0(Z)
		=
		0.
	\end{equation*}
	
	Moreover, by the partially linear model:
	\begin{equation*}
		Y=\theta_0 T + f(X)+\zeta,
		\mathbb{E}[\zeta\mid X,T]=0.
	\end{equation*}
	Since $Z=\Phi(X)$ is a function of $X$, the tower property gives
	\begin{equation*}
		\mathbb{E}[\zeta\mid Z,T]
		=
		\mathbb{E}\!\left[\mathbb{E}[\zeta\mid X,T]\mid Z,T\right]
		=
		0.
	\end{equation*}
	
	At the true parameter pair $(\theta_0,\eta_0)$, define
	\begin{equation*}
		u:=Y-g_0(Z)-\theta_0\big(T-m_0(Z)\big).
	\end{equation*}
	
	Using the PLM and the identity
	\begin{equation*}
		g_0(Z)=\mathbb{E}[Y\mid Z]
		=
		\theta_0 m_0(Z)+\mathbb{E}[f(X)\mid Z],
	\end{equation*}
	we obtain
	\begin{equation*}
		u=f(X)-\mathbb{E}[f(X)\mid Z]+\zeta.
	\end{equation*}
	
	Hence,
	\begin{equation*}
		\mathbb{E}[u\mid Z]=0,
	\end{equation*}
	because
	\[
	\mathbb{E}\!\left[f(X)-\mathbb{E}[f(X)\mid Z]\mid Z\right]=0
	\]
	and
	\[
	\mathbb{E}[\zeta\mid Z]=0.
	\]
	
	Now consider a perturbation path $\eta_t=(g_t,m_t)$ with
	\[
	g_t=g_0+t h_g,
	m_t=m_0+t h_m,
	\]
	where $h_g$ and $h_m$ are measurable perturbation functions defined on the corresponding representation subspaces. We compute the pathwise derivative of the moment at $(\theta_0,\eta_t)$:
	{\small
		\begin{align*}
			\left.\frac{d}{dt}\,
			\mathbb{E}\!\left[\psi_Z(W;\theta_0,\eta_t)\right]\right|_{t=0}
			&=
			\mathbb{E}\!\Big[\big(-h_g(Z)+\theta_0 h_m(Z)\big)v-h_m(Z)u\Big].
	\end{align*}}
	
	Since $h_g(Z)$ and $h_m(Z)$ are $Z$-measurable, iterated expectations yield
	\begin{align*}
		\mathbb{E}[h_g(Z)\,v]
		&=
		\mathbb{E}\!\left[h_g(Z)\,\mathbb{E}[v\mid Z]\right]
		=
		0,\\
		\mathbb{E}[h_m(Z)\,v]
		&=
		\mathbb{E}\!\left[h_m(Z)\,\mathbb{E}[v\mid Z]\right]
		=
		0,\\
		\mathbb{E}[h_m(Z)\,u]
		&=
		\mathbb{E}\!\left[h_m(Z)\,\mathbb{E}[u\mid Z]\right]
		=
		0.
	\end{align*}
	
	Therefore,
	\begin{equation*}
		\left.\frac{d}{dt}\,
		\mathbb{E}\!\left[\psi_Z(W;\theta_0,\eta_t)\right]\right|_{t=0}
		=
		0,
	\end{equation*}
	which is exactly the Neyman orthogonality condition:
	\begin{equation*}
		\left.
		\frac{\partial}{\partial \eta}
		\mathbb{E}\!\left[\psi_Z(W;\theta_0,\eta)\right]
		\right|_{\eta=\eta_0}
		=
		0.
	\end{equation*}
\end{proof}

Theorem 5.1 implies that DDML is locally insensitive to small perturbations in the nuisance functions, so that first-order nuisance estimation errors do not affect the causal effect estimation.  Assumption 5.3 formalizes the role of RDO by ensuring that the interaction between nuisance estimation errors is asymptotically negligible. Building on this, Theorem 5.2 shows consistency of the cross-fitted DDML estimator.

\noindent \textbf{Theorem 5.2.}
\textit{Under Assumptions 5.1--5.3, let $\hat{\theta}$ be the DDML estimator obtained from the empirical orthogonal moment equation with cross-fitted nuisance predictions. Then}
\begin{equation}
	\hat{\theta} \xrightarrow{p} \theta_0 \quad
	\textit{as } \quad N\to\infty.
	\label{eq:ddml_consistency}
\end{equation}
\begin{proof}
	
	Define the cross-fitted residuals
	\begin{equation*}
		\hat v_i:=T_i-\hat m(Z_i),	
		\hat u_i:=Y_i-\hat g(Z_i),
	\end{equation*}
	where $m(Z)=m(Z_c,Z_t)$ and $g(Z)=g(Z_c,Z_y)$.
	Set
	\begin{equation*}
		\hat Q:=\frac{1}{N}\sum_{i=1}^N \hat v_i\hat u_i,
		\hat D:=\frac{1}{N}\sum_{i=1}^N \hat v_i^2,
		\hat\theta=\frac{\hat Q}{\hat D}.
	\end{equation*}
	
	Let
	\begin{equation*}
		v_i^0:=T_i-m(Z_i),
		u_i^0:=Y_i-g(Z_i),
	\end{equation*}
	and denote
	\begin{equation*}
		Q:=\mathbb{E}[v^0u^0],
		D:=\mathbb{E}\!\left[(v^0)^2\right].
	\end{equation*}
	
	\paragraph{(i) Denominator.}
	Write
	\begin{equation*}
		\hat v_i=v_i^0+\Delta_{m,i},
		\Delta_{m,i}:=m(Z_i)-\hat m(Z_i).
	\end{equation*}
	
	Then
	\begin{equation*}
		\hat v_i^2-(v_i^0)^2
		=
		2v_i^0\Delta_{m,i}+\Delta_{m,i}^2.
	\end{equation*}
	
	Consequently, by the triangle inequality and Cauchy--Schwarz,
	\begin{equation*}
		\begin{aligned}
			\Big|\hat D-\frac{1}{N}\sum_{i=1}^N (v_i^0)^2\Big|
			&\le
			\frac{2}{N}\sum_{i=1}^N |v_i^0\Delta_{m,i}|
			+\frac{1}{N}\sum_{i=1}^N \Delta_{m,i}^2 \\
			&\le
			2\Big(\frac{1}{N}\sum_{i=1}^N (v_i^0)^2\Big)^{1/2}
			\Big(\frac{1}{N}\sum_{i=1}^N \Delta_{m,i}^2\Big)^{1/2} \\
			&\quad+
			\frac{1}{N}\sum_{i=1}^N \Delta_{m,i}^2 .
		\end{aligned}
	\end{equation*}
	
	By Assumption 5.2 and the law of large numbers,
	\begin{equation*}
		\frac{1}{N}\sum_{i=1}^N (v_i^0)^2 \xrightarrow{p} D,
	\end{equation*}
	and by Assumption 5.3,
	\begin{equation*}
		\frac{1}{N}\sum_{i=1}^N \Delta_{m,i}^2
		=
		\|\hat m-m\|_{L_2}^2
		=
		o_p(1).
	\end{equation*}
	
	Thus,
	\begin{equation*}
		\hat D\xrightarrow{p}D.
	\end{equation*}
	
	Moreover, Assumption 5.2 implies that $D\ge c>0$, hence
	\begin{equation*}
		\Pr(\hat D>c/2)\to 1.
	\end{equation*}
	
	\paragraph{(ii) Numerator.}
	Write
	\begin{equation*}
		\hat u_i=u_i^0+\Delta_{g,i},
		\Delta_{g,i}:=g(Z_i)-\hat g(Z_i).
	\end{equation*}
	
	Then
	\begin{equation*}
		\hat v_i\hat u_i-v_i^0u_i^0
		=
		v_i^0\Delta_{g,i}+u_i^0\Delta_{m,i}+\Delta_{m,i}\Delta_{g,i}.
	\end{equation*}
	
	Hence, by Cauchy--Schwarz,
	\begin{equation*}
		\begin{aligned}
			\Big|\hat Q-\frac{1}{N}\sum_{i=1}^N v_i^0u_i^0\Big|
			&\le
			\Big(\frac{1}{N}\sum_{i=1}^N (v_i^0)^2\Big)^{1/2}
			\Big(\frac{1}{N}\sum_{i=1}^N \Delta_{g,i}^2\Big)^{1/2} \\
			&\quad+
			\Big(\frac{1}{N}\sum_{i=1}^N (u_i^0)^2\Big)^{1/2}
			\Big(\frac{1}{N}\sum_{i=1}^N \Delta_{m,i}^2\Big)^{1/2} \\
			&\quad+
			\Big(\frac{1}{N}\sum_{i=1}^N \Delta_{m,i}^2\Big)^{1/2}
			\Big(\frac{1}{N}\sum_{i=1}^N \Delta_{g,i}^2\Big)^{1/2}.
		\end{aligned}
	\end{equation*}
	
	By Assumption 5.2,
	\begin{equation*}
		\frac{1}{N}\sum_{i=1}^N (u_i^0)^2=o_p(1),
		\frac{1}{N}\sum_{i=1}^N (v_i^0)^2=o_p(1),
	\end{equation*}
	while Assumption 5.3 gives
	\begin{equation*}
		\frac{1}{N}\sum_{i=1}^N \Delta_{m,i}^2=o_p(1),
		\frac{1}{N}\sum_{i=1}^N \Delta_{g,i}^2=o_p(1).
	\end{equation*}
	
	Therefore,
	\begin{equation*}
		\hat Q-\frac{1}{N}\sum_{i=1}^N v_i^0u_i^0=o_p(1).
	\end{equation*}
	
	By the law of large numbers,
	\begin{equation*}
		\frac{1}{N}\sum_{i=1}^N v_i^0u_i^0 \xrightarrow{p} Q,
	\end{equation*}
	and hence
	\begin{equation*}
		\hat Q\xrightarrow{p}Q.
	\end{equation*}
	
	\paragraph{(iii) Identify the limit.}
	Under the partially linear model and Assumption 5.1,
	\begin{equation*}
		u^0
		=
		Y-g(Z)
		=
		\theta_0\big(T-m(Z)\big)+\zeta
		=
		\theta_0 v^0+\zeta.
	\end{equation*}
	
	Thus,
	\begin{equation*}
		Q
		=
		\mathbb{E}[v^0u^0]
		=
		\theta_0\,\mathbb{E}\!\left[(v^0)^2\right]+\mathbb{E}[v^0\zeta]
		=
		\theta_0 D+\mathbb{E}[v^0\zeta].
	\end{equation*}
	
	Since
	\begin{equation*}
		\mathbb{E}[\zeta\mid Z,T]
		=
		\mathbb{E}\!\left[\mathbb{E}[\zeta\mid X,T]\mid Z,T\right]
		=
		0,
	\end{equation*}
	we have
	\begin{equation*}
		\mathbb{E}[v^0\zeta]
		=
		\mathbb{E}\!\left[v^0\,\mathbb{E}[\zeta\mid Z,T]\right]
		=
		0.
	\end{equation*}
	
	Hence,
	\begin{equation*}
		Q=\theta_0 D.
	\end{equation*}
	
	Combining (i)--(iii), we have
	\begin{equation*}
		(\hat Q,\hat D)\xrightarrow{p}(Q,D),
		D>0.
	\end{equation*}
	
	By Slutsky's theorem,
	\begin{equation*}
		\hat\theta
		=
		\frac{\hat Q}{\hat D}
		\xrightarrow{p}
		\frac{Q}{D}
		=
		\theta_0.
	\end{equation*}
\end{proof}

These analyses provide the theoretical foundation of DDML by showing that CRD preserves the target causal effect and RDO supports consistent effect estimation under cross-fitting.

\section{Experiments}\label{Ex}
\subsection{Experiment Settings}
\subsubsection{Datasets}
We used the following synthetic, semi-synthetic, and real-world datasets to comprehensively evaluate the effectiveness of the proposed method.

\textbullet{\textbf{Synthetic data.}} We generated synthetic datasets with both \emph{binary} and \emph{continuous} treatments to evaluate the effectiveness of our method under strong nonlinear confounding and highly mixed covariates. For each setting, we sampled $N=6000$ instances with feature dimensions $d\in\{20,50,100,200\}$ and repeated each $(N,d)$ configuration over 20 independent runs. Across all settings, the true causal effect is fixed to $\theta=5.0$.

Following the settings in \cite{ hassanpour2019learning,wu2023learning}, in each experiment, we first draw latent variables $Z\in\mathbb{R}^{N\times d}$ with i.i.d.\ entries $Z_{ij}\sim\mathcal{N}(0,1)$.
To induce strong entanglement in the observed space, we sample a random mixing matrix $M\in\mathbb{R}^{d\times d}$ with entries $M_{ij}\sim\mathcal{N}(0,\lambda^2)$, where $\lambda=8.0$ controls the entanglement strength.
We then construct a linearly mixed representation:
\begin{equation}
	X_{\mathrm{lin}} = Z + \frac{1}{d} ZM,
\end{equation}
and further apply nonlinear distortions:
\begin{equation}
	X = \tanh\!\bigl(X_{\mathrm{lin}}\bigr)
	+ 0.2\,\sin\!\left(\frac{X_{\mathrm{lin}}M^\top}{d}\right),
\end{equation}
yielding the observed covariates $X\in\mathbb{R}^{N\times d}$.
This procedure globally mixes coordinates and introduces strong nonlinearity, producing highly mixed and non-axis-aligned representations.

Given $X$, we first define a shared nonlinear confounder $Z_c$, which affects both treatment assignment and outcome generation:
\begin{equation}
	Z_c = 0.6\,X_{:,0}\odot X_{:,1}
	+ 0.4\,X_{:,2}^{\odot 2}
	+ 0.3\,\sin\!\bigl(X_{:,3}+X_{:,4}\bigr),
\end{equation}
where $\odot$ denotes elementwise product and $X_{:,k}$ denotes the $k$-th feature column.
We further construct a treatment-specific factor $Z_t$, which affects treatment assignment but does not directly enter outcome generation:
\begin{equation}
	Z_t = 0.5\,X_{:,5}\odot X_{:,6}
	+ 0.3\,\tanh\!\bigl(X_{:,7}\bigr)
	+ 0.2\,\cos\!\bigl(X_{:,8}+X_{:,9}\bigr).
\end{equation}
In addition, we define an outcome-specific factor $Z_y$, which affects the outcome but does not directly drive treatment assignment:
\begin{equation}
	Z_y = 0.5\,X_{:,1}\odot X_{:,2}
	+ 0.3\,\cos\!\bigl(X_{:,0}+X_{:,3}\bigr).
\end{equation}
The treatment $T$ is then generated in two variants while keeping $(X,Z_c,Z_t,Z_y)$ fixed, so that the binary- and continuous-treatment settings differ only in the treatment assignment mechanism.

For the \emph{binary-treatment} setting, we generate $T\in\{0,1\}$ according to
\begin{equation}
	\Pr(T_i=1\mid X_i)
	=
	\sigma\!\bigl(\alpha_c Z_{c,i}+\alpha_t Z_{t,i}\bigr),
	\sigma(u)=\frac{1}{1+e^{-u}},
\end{equation}
where $\alpha_c=4.0$ controls the confounding strength and $\alpha_t=2.0$ controls the treatment-specific signal strength. We then sample
\begin{equation}
	T_i \sim \mathrm{Bernoulli}\!\left(\Pr(T_i=1\mid X_i)\right).
\end{equation}
This induces strong dependence between $T$ and $X$ through nonlinear interactions, making naive treated--control comparisons biased.

For the \emph{continuous-treatment} setting, we generate a noisy nonlinear treatment signal from both the shared confounder and the treatment-specific factor:
\begin{equation}
	T_i
	=
	\alpha_c\Bigl(Z_{c,i}+0.5\,\tanh(Z_{c,i})\Bigr)
	+
	\alpha_t\Bigl(Z_{t,i}+0.3\,\sin(Z_{t,i})\Bigr)
	+
	\eta_i,
\end{equation}
where $\alpha_c=4.0$, $\alpha_t=2.0$, and $\eta_i\sim\mathcal{N}(0,\sigma_T^2)$ with $\sigma_T=1.0$.
Unless otherwise stated, we further standardize $T$ to have zero mean and unit variance to stabilize learning across different feature dimensions.

Finally, for \emph{both} treatment variants, outcomes are generated as
\begin{equation}
	Y_i = \alpha\,(Z_{c,i}+Z_{y,i}) + \theta\,T_i + \varepsilon_i,
\end{equation}
where $\varepsilon_i\sim\mathcal{N}(0,\sigma_Y^2)$ with $\sigma_Y=1.0$.
Here, $Z_t$ affects the outcome only indirectly through $T_i$ and does not directly appear in the outcome equation.

\textbullet{\textbf{Semi-synthetic data.}} \textit{IHDP} (Infant Health and Development Program) is derived from a real randomized controlled trial that investigated the effect of professional home-visiting services on children's cognitive development. We use the widely adopted semi-synthetic benchmark introduced in \cite{shalit2017estimating}, which contains 747 samples (139 treated and 608 control), each with 25 covariates. Following prior work, we report results averaged over 100 replications. \textit{Twins} is a semi-synthetic dataset designed to study the causal relationship between birth-related factors and infant mortality outcomes. We adopt the experimental setup in \cite{louizos2017causal}, where each sample is represented by 46 covariates. To reduce variance and ensure stable comparison, we report results averaged over 10 replications.

\textbullet{\textbf{Real-world data.}} \textit{Jobs} is a widely used dataset for benchmarking causal effect estimation in observational studies, aiming to quantify the effect of participating in a job-training program on individuals' income. Following previous work \cite{shalit2017estimating}, we report the mean performance over 10 independent replications.

\subsubsection{Comparison Algorithms} We compared the proposed DDML with 13 representative baseline methods from the following categories. \textbf{Traditional methods}: \textbf{OLS-1} treats the treatment as a standard feature and estimates the effect via a single linear regression; \textbf{OLS-2} fits separate regressions for the treated and control groups; \textbf{KNN} infers counterfactual outcomes using cross-group nearest neighbors; \textbf{BART} captures nonlinear relationships with a nonparametric tree-ensemble model. \textbf{Representation learning methods}: \textbf{CFR-wass/CFR-mmd} \cite{shalit2017estimating} learn balanced representations across groups via Wasserstein and MMD regularization; \textbf{SITE} \cite{yao2018representation} preserves local similarity among samples under global balance; \textbf{DeR-CFR} \cite{wu2023learning} reduces selection bias by decoupling representations to separate confounding and adjustment information; \textbf{DIGNet} \cite{huang2024dignet} enhances robustness by decoupling latent patterns and imposing geometric constraints; \textbf{FCCL} \cite{zhangcounterfactual} combines normalizing flows with counterfactual contrastive learning to generate and align counterfactual representations. \textbf{Debiased estimation methods}: \textbf{DML-RF/DML-MLP} \cite{chernozhukov2018double} estimate nuisance functions using random forest and multi-layer perceptron within DML; \textbf{C-DML} \cite{fingerhut2022coordinated} employs coordinated nuisance training to estimate the causal effect.

\subsubsection{Evaluation Metrics}
For the semi-synthetic dataset, \textit{IHDP} and \textit{Twins}, the ground-truth causal effect is available. Following prior work \cite{kuang2017treatment}, we evaluate methods using the mean absolute error (MAE) and root mean squared error (RMSE) of the estimated average treatment effect (ATE) over repeated trials. Specifically, given $R$ repeated experiments, let {\footnotesize$\hat{ATE}^{(r)}$} denote the estimated ATE in the $r$-th run and {\footnotesize$ATE^\star$} denote the true ATE. We compute {\footnotesize $\mathrm{MAE}_{\mathrm{ATE}}=\frac{1}{R}\sum_{r=1}^{R}\left|\hat{ATE}^{(r)}-ATE^\star\right|$},
{\footnotesize$\mathrm{RMSE}_{\mathrm{ATE}}=\sqrt{\frac{1}{R}\sum_{r=1}^{R}\left(\hat{ATE}^{(r)}-ATE^\star\right)^2 }.$} For the \textit{Jobs} dataset, we similarly report MAE and RMSE of the estimated average treatment effect on the treated (ATT) over repeated trials. Let {\footnotesize$\hat{ATT}^{(r)}$} be the estimated ATT in the $r$-th run and {\footnotesize$ATT^\star$} be the target ATT. The metrics are defined as {\footnotesize$\mathrm{MAE}_{\mathrm{ATT}}=\frac{1}{R}\sum_{r=1}^{R}\left|\hat{ATT}^{(r)}-ATT^\star\right|$},
{\footnotesize$\mathrm{RMSE}_{\mathrm{ATT}}=\sqrt{\frac{1}{R}\sum_{r=1}^{R}\left(\hat{ATT}^{(r)}-ATT^\star\right)^2 }.$ }

\subsubsection{Implementation Details} All experiments were conducted on a Windows 10 computer equipped with an Intel(R) Core(TM) i9-10900F 2.80GHz CPU, NVIDIA GeForce RTX 3060 GPU, and 32GB RAM. All algorithms were implemented in \textit{Python}. For both \textsc{DML} and \textsc{DDML}, we adopt 5-fold cross-fitting ($K=5$) for data splitting and effect estimation, employing both MLP and RF as nuisance estimators.
\begin{table*}[t]
	\centering
	\caption{Performance comparison under different feature dimensions (\textit{Binary\_20}, \textit{Binary\_50}, \textit{Binary\_100}, \textit{Binary\_200}). Results are reported as mean $\pm$ std (Best results are highlighted in bold, and second-best results are underlined.)}
	\setlength{\tabcolsep}{3pt}
	\renewcommand{\arraystretch}{1.2}
	\label{tab:bi_data}
	\resizebox{\textwidth}{!}{
		\begin{tabular}{l|cccccccc}
			\toprule
			& \multicolumn{2}{c}{Binary\_20} & \multicolumn{2}{c}{Binary\_50} & \multicolumn{2}{c}{Binary\_100} & \multicolumn{2}{c}{Binary\_200} \\
			\midrule
			Method & MAE & RMSE & MAE & RMSE & MAE & RMSE & MAE & RMSE \\
			\midrule
			OLS-1 & 1.9949$\pm${\scriptsize0.4142} & 2.0353$\pm${\scriptsize0.4142} & 1.7467$\pm${\scriptsize0.1638} & 1.7540$\pm${\scriptsize0.1638} & 1.8085$\pm${\scriptsize0.1771} & 1.8167$\pm${\scriptsize0.1771} & 1.4796$\pm${\scriptsize0.0942} & 1.4825$\pm${\scriptsize0.0942} \\
			OLS-2 & 2.0077$\pm${\scriptsize0.4142} & 2.0479$\pm${\scriptsize0.4142} & 1.7528$\pm${\scriptsize0.1659} & 1.7602$\pm${\scriptsize0.1659} & 1.8113$\pm${\scriptsize0.1745} & 1.8193$\pm${\scriptsize0.1745} & 1.4846$\pm${\scriptsize0.0959} & 1.4875$\pm${\scriptsize0.0959} \\
			KNN & 1.6674$\pm${\scriptsize0.2652} & 1.6873$\pm${\scriptsize0.2652} & 1.7733$\pm${\scriptsize0.1331} & 1.7780$\pm${\scriptsize0.1331} & 1.9194$\pm${\scriptsize0.1437} & 1.9245$\pm${\scriptsize0.1437} & 1.6944$\pm${\scriptsize0.0864} & 1.6965$\pm${\scriptsize0.0864} \\
			BART & 0.9369$\pm${\scriptsize0.1938} & 0.9558$\pm${\scriptsize0.1938} & 0.8018$\pm${\scriptsize0.1290} & 0.8116$\pm${\scriptsize0.1290} & 0.8231$\pm${\scriptsize0.1090} & 0.8299$\pm${\scriptsize0.1090} & 0.6041$\pm${\scriptsize0.0825} & 0.6094$\pm${\scriptsize0.0825} \\
			\midrule
			CFR-wass & 0.2005$\pm${\scriptsize0.1311} & 0.2377$\pm${\scriptsize0.1311} & 0.1936$\pm${\scriptsize0.1227} & 0.2276$\pm${\scriptsize0.1227} & 0.3556$\pm${\scriptsize0.1543} & 0.3861$\pm${\scriptsize0.1543} & 0.3697$\pm${\scriptsize0.1713} & 0.4057$\pm${\scriptsize0.1713} \\
			CFR-mmd & 0.2018$\pm${\scriptsize0.1372} & 0.2421$\pm${\scriptsize0.1372} & 0.3092$\pm${\scriptsize0.1468} & 0.3407$\pm${\scriptsize0.1468} & 0.3609$\pm${\scriptsize0.1228} & 0.3802$\pm${\scriptsize0.1228} & 0.5191$\pm${\scriptsize0.1147} & 0.5310$\pm${\scriptsize0.1147} \\
			SITE & 0.2577$\pm${\scriptsize0.1346} & 0.2892$\pm${\scriptsize0.1346} & 0.4150$\pm${\scriptsize0.1499} & 0.4400$\pm${\scriptsize0.1499} & 0.6768$\pm${\scriptsize0.1498} & 0.6923$\pm${\scriptsize0.1498} & 0.8667$\pm${\scriptsize0.2050} & 0.8895$\pm${\scriptsize0.2050} \\
			DeR-CFR & 0.1584$\pm${\scriptsize0.0707} & 0.1727$\pm${\scriptsize0.0707} & 0.1369$\pm${\scriptsize0.0952} & 0.1654$\pm${\scriptsize0.0952} & \textbf{0.1432$\pm${\scriptsize0.1263}} & \underline{0.1888$\pm${\scriptsize0.1263}} & 0.2876$\pm${\scriptsize0.1214} & 0.3219$\pm${\scriptsize0.1214} \\
			DIGNet & 0.1949$\pm${\scriptsize0.1000} & 0.2180$\pm${\scriptsize0.1000} & 0.2595$\pm${\scriptsize0.1679} & 0.3068$\pm${\scriptsize0.1679} & 0.2750$\pm${\scriptsize0.1472} & 0.3102$\pm${\scriptsize0.1472} & 0.3784$\pm${\scriptsize0.1757} & 0.4154$\pm${\scriptsize0.1757} \\
			FCCL & 0.1200$\pm${\scriptsize0.0981} & 0.1534$\pm${\scriptsize0.0981} & 0.1748$\pm${\scriptsize0.1311} & 0.2166$\pm${\scriptsize0.1311} & 0.2611$\pm${\scriptsize0.1073} & 0.2812$\pm${\scriptsize0.1073} & 0.2999$\pm${\scriptsize0.1022} & 0.3160$\pm${\scriptsize0.1022} \\
			\midrule
			DML-RF & 1.9373$\pm${\scriptsize0.3596} & 1.9687$\pm${\scriptsize0.3596} & 1.7243$\pm${\scriptsize0.1647} & 1.7318$\pm${\scriptsize0.1647} & 1.7708$\pm${\scriptsize0.1650} & 1.7781$\pm${\scriptsize0.1650} & 1.4772$\pm${\scriptsize0.0896} & 1.4798$\pm${\scriptsize0.0896} \\
			DML-MLP & 0.4292$\pm${\scriptsize0.0706} & 0.4347$\pm${\scriptsize0.0706} & 0.7790$\pm${\scriptsize0.0923} & 0.7841$\pm${\scriptsize0.0923} & 1.0111$\pm${\scriptsize0.1056} & 1.0163$\pm${\scriptsize0.1056} & 0.5759$\pm${\scriptsize0.1748} & 0.6005$\pm${\scriptsize0.1748} \\
			C-DML & 0.3755$\pm${\scriptsize0.2795} & 0.4639$\pm${\scriptsize0.2795} & 0.7999$\pm${\scriptsize0.2375} & 0.8327$\pm${\scriptsize0.2375} & 1.0634$\pm${\scriptsize0.2778} & 1.0974$\pm${\scriptsize0.2778} & 1.1308$\pm${\scriptsize0.3710} & 1.1872$\pm${\scriptsize0.3710} \\
			\midrule
			\textbf{DDML-RF} & \underline{0.0769$\pm${\scriptsize0.0522}} & \underline{0.0922$\pm${\scriptsize0.0522}} & \textbf{0.0520$\pm${\scriptsize0.0337}} & \textbf{0.0615$\pm${\scriptsize0.0337}} & \underline{0.1550$\pm${\scriptsize0.0911}} & \textbf{0.1786$\pm${\scriptsize0.0911}} & \underline{0.2615$\pm${\scriptsize0.0862}} & \underline{0.2747$\pm${\scriptsize0.0862}} \\
			\textbf{DDML-MLP} & \textbf{0.0488$\pm${\scriptsize0.0295}} & \textbf{0.0567$\pm${\scriptsize0.0295}} & \underline{0.1278$\pm${\scriptsize0.0610}} & \underline{0.1409$\pm${\scriptsize0.0610}} & 0.2379$\pm${\scriptsize0.0727} &
			0.2452$\pm${\scriptsize0.0727} & \textbf{0.1942$\pm${\scriptsize0.1045}} & \textbf{0.2193$\pm${\scriptsize0.1045}} \\
			\bottomrule
	\end{tabular}}\\[4pt]
	\begin{minipage}{\linewidth}
		\footnotesize\raggedright
		{\footnotesize **DDML achieves average improvements of \textbf{36.4\%} in MAE and \textbf{40.5\%} in RMSE compared with the best baseline method.}
	\end{minipage}
\end{table*}

\begin{table*}[t]
	\centering
	\caption{Performance comparison under different feature dimensions (\textit{Continuous\_20}, \textit{Continuous\_50}, \textit{Continuous\_100}, \textit{Continuous\_200}). Results are reported as mean $\pm$ std (Best results are highlighted in bold, and second-best results are underlined.)}
	\label{tab:con_data}
	\setlength{\tabcolsep}{3pt}
	\renewcommand{\arraystretch}{1.2}
	\resizebox{\textwidth}{!}{
		\begin{tabular}{l|cccccccc}
			\toprule
			& \multicolumn{2}{c}{Continuous\_20} & \multicolumn{2}{c}{Continuous\_50} & \multicolumn{2}{c}{Continuous\_100} & \multicolumn{2}{c}{Continuous\_200} \\
			
			\midrule
			Method & MAE & RMSE & MAE & RMSE & MAE & RMSE & MAE & RMSE \\
			
			\midrule
			OLS-1
			& 1.6857$\pm${\scriptsize0.2160} & 1.6988$\pm${\scriptsize0.2160}
			& 1.5734$\pm${\scriptsize0.1384} & 1.5791$\pm${\scriptsize0.1384}
			& 1.5773$\pm${\scriptsize0.1054} & 1.5806$\pm${\scriptsize0.1054}
			& 1.3991$\pm${\scriptsize0.0679} & 1.4006$\pm${\scriptsize0.0679} \\
			
			BART
			& 1.1659$\pm${\scriptsize0.4326} & 1.2398$\pm${\scriptsize0.4326}
			& 1.0551$\pm${\scriptsize0.3297} & 1.1030$\pm${\scriptsize0.3297}
			& 0.9282$\pm${\scriptsize0.3812} & 0.9998$\pm${\scriptsize0.3812}
			& 0.8684$\pm${\scriptsize0.3149} & 0.9210$\pm${\scriptsize0.3149} \\
			
			\midrule
			DML-RF
			& 1.6517$\pm${\scriptsize0.1976} & 1.6629$\pm${\scriptsize0.1976}
			& 1.5647$\pm${\scriptsize0.1304} & 1.5699$\pm${\scriptsize0.1304}
			& 1.5649$\pm${\scriptsize0.1051} & 1.5683$\pm${\scriptsize0.1051}
			& 1.3926$\pm${\scriptsize0.0663} & 1.3941$\pm${\scriptsize0.0663} \\
			
			DML-MLP
			& 1.5191$\pm${\scriptsize0.0793} & 1.5248$\pm${\scriptsize0.0793}
			& 1.1625$\pm${\scriptsize0.0831} & 1.1653$\pm${\scriptsize0.0831}
			& 1.5213$\pm${\scriptsize0.1102} & 1.5251$\pm${\scriptsize0.1102}
			& 1.3159$\pm${\scriptsize0.1102} & 1.3202$\pm${\scriptsize0.1102} \\
			
			C-DML
			& 0.4039$\pm${\scriptsize0.6964} & 0.7898$\pm${\scriptsize0.6964}
			& 0.7852$\pm${\scriptsize0.1230} & 0.7943$\pm${\scriptsize0.1230}
			& 1.2910$\pm${\scriptsize0.1603} & 1.3004$\pm${\scriptsize0.1603}
			& 0.7457$\pm${\scriptsize0.1888} & 0.7680$\pm${\scriptsize0.1888} \\
			
			\midrule
			\textbf{DDML-RF}
			& \underline{0.2400$\pm${\scriptsize0.1011}} & \underline{0.2595$\pm${\scriptsize0.1011}}
			& \textbf{0.0706$\pm${\scriptsize0.0425}} & \textbf{0.0819$\pm${\scriptsize0.0425}}
			& \underline{0.3943$\pm${\scriptsize0.3035}} & \underline{0.4930$\pm${\scriptsize0.3035}}
			& \textbf{0.2455$\pm${\scriptsize0.1900}} & \textbf{0.3075$\pm${\scriptsize0.1900}} \\
			
			\textbf{DDML-MLP}
			& \textbf{0.0820$\pm${\scriptsize0.0649}} & \textbf{0.1036$\pm${\scriptsize0.0649}}
			& \underline{0.2416$\pm${\scriptsize0.0465}}& \underline{0.2459$\pm${\scriptsize0.0465}}
			& \textbf{0.2226$\pm${\scriptsize0.1731}} & \textbf{0.2793$\pm${\scriptsize0.1731}}
			& \underline{0.4519$\pm${\scriptsize0.1543}} & \underline{0.4763$\pm${\scriptsize0.1543}} \\
			\bottomrule
	\end{tabular}}\\[4pt]
	\begin{minipage}{\linewidth}
		\footnotesize\raggedright
		{\footnotesize **DDML achieves average improvements of \textbf{78.5\%} in MAE and \textbf{77.2\%} in RMSE compared with the best baseline method.}
	\end{minipage}
\end{table*}

\begin{table*}[t]
	\centering
	\caption{Performance comparison on \textit{IHDP}, \textit{Twins} and \textit{Jobs} datasets. Results are reported as mean $\pm$ std. (Best results are highlighted in bold, and second-best results are underlined.)}
	\label{tab:real_results}
	\setlength{\tabcolsep}{5pt}
	\renewcommand{\arraystretch}{1.1}
	\resizebox{0.9\textwidth}{!}{
		\begin{tabular}{l|cccccc}
			\toprule
			& \multicolumn{2}{c}{IHDP} & \multicolumn{2}{c}{Twins} & \multicolumn{2}{c}{Jobs} \\
			\midrule
			Method & MAE & RMSE & MAE & RMSE & MAE & RMSE \\
			\midrule
			OLS-1
			& 0.7474$\pm${\scriptsize1.4530} & 1.6275$\pm${\scriptsize1.4530}
			& 0.0016$\pm${\scriptsize0.0009} & 0.0018$\pm${\scriptsize0.0009}
			& 0.1280$\pm${\scriptsize0.0000} & 0.1280$\pm${\scriptsize0.0000} \\
			OLS-2
			& \underline{0.1340$\pm${\scriptsize0.1134}} & \textbf{0.1752$\pm${\scriptsize0.1134}}
			& 0.0013$\pm${\scriptsize0.0011} & 0.0016$\pm${\scriptsize0.0011}
			& 0.0385$\pm${\scriptsize0.0000} & 0.0385$\pm${\scriptsize0.0000} \\
			KNN
			& 0.1513$\pm${\scriptsize0.1790} & 0.2337$\pm${\scriptsize0.1790}
			& 0.0016$\pm${\scriptsize0.0015} & 0.0022$\pm${\scriptsize0.0015}
			& 0.1288$\pm${\scriptsize0.0001} & 0.1288$\pm${\scriptsize0.0001} \\
			BART
			& 0.4036$\pm${\scriptsize0.7487} & 0.8472$\pm${\scriptsize0.7487}
			& 0.0026$\pm${\scriptsize0.0016} & 0.0030$\pm${\scriptsize0.0016}
			& 0.0652$\pm${\scriptsize0.0217} & 0.0684$\pm${\scriptsize0.0217} \\
			
			\midrule
			CFR-mmd
			& 0.2852$\pm${\scriptsize0.1262} & 0.2882$\pm${\scriptsize0.1262}
			& 0.0059$\pm${\scriptsize0.0028} & 0.0059$\pm${\scriptsize0.0028}
			& 0.1583$\pm${\scriptsize0.0727} & 0.1727$\pm${\scriptsize0.0727} \\
			CFR-wass
			& 0.2153$\pm${\scriptsize0.1721} & 0.2702$\pm${\scriptsize0.1721}
			& 0.0036$\pm${\scriptsize0.0028} & 0.0036$\pm${\scriptsize0.0028}
			& 0.1574$\pm${\scriptsize0.0368} & 0.1612$\pm${\scriptsize0.0368} \\
			SITE
			& 0.5365$\pm${\scriptsize0.9607} & 1.0576$\pm${\scriptsize0.9607}
			& 0.0027$\pm${\scriptsize0.0017} & 0.0027$\pm${\scriptsize0.0017}
			& 0.1173$\pm${\scriptsize0.0280} & 0.1203$\pm${\scriptsize0.0280} \\
			DeR-CFR
			& 0.2602$\pm${\scriptsize0.2742} & 0.3680$\pm${\scriptsize0.2742}
			& 0.0035$\pm${\scriptsize0.0038} & 0.0035$\pm${\scriptsize0.0038}
			& 0.0768$\pm${\scriptsize0.0532} & 0.0919$\pm${\scriptsize0.0532} \\
			DIGNet
			& 0.2028$\pm${\scriptsize0.1936} & 0.2736$\pm${\scriptsize0.1936}
			& 0.0043$\pm${\scriptsize0.0029} & 0.0043$\pm${\scriptsize0.0029}
			& 0.1066$\pm${\scriptsize0.0508} & 0.1170$\pm${\scriptsize0.0508} \\
			FCCL
			& 0.2175$\pm${\scriptsize0.2192} & 0.3009$\pm${\scriptsize0.2192}
			& 0.0051$\pm${\scriptsize0.0034} & 0.0051$\pm${\scriptsize0.0034}
			& 0.0788$\pm${\scriptsize0.0551} & 0.0946$\pm${\scriptsize0.0551} \\
			
			\midrule
			DML-RF
			& 0.7253$\pm${\scriptsize1.7363} & 1.7998$\pm${\scriptsize1.7363}
			& 0.0081$\pm${\scriptsize0.0090} & 0.0082$\pm${\scriptsize0.0073}
			& 0.0524$\pm${\scriptsize0.0333} & 0.0612$\pm${\scriptsize0.0333} \\
			DML-MLP
			& 0.9696$\pm${\scriptsize1.5547} & 1.7651$\pm${\scriptsize1.5547}
			& 0.2192$\pm${\scriptsize0.2851} & 0.2518$\pm${\scriptsize0.2225}
			& 0.0573$\pm${\scriptsize0.0302} & 0.0641$\pm${\scriptsize0.0302} \\
			C-DML
			& 3.4081$\pm${\scriptsize2.4953} & 4.1496$\pm${\scriptsize2.4953}
			& 0.0954$\pm${\scriptsize0.2606} & 0.0954$\pm${\scriptsize0.2606}
			& 0.0737$\pm${\scriptsize0.0122} & 0.0746$\pm${\scriptsize0.0122} \\
			
			\midrule
			\textbf{DDML-RF}
			& 0.1450$\pm${\scriptsize0.1506} & 0.2035$\pm${\scriptsize0.1506}
			& \textbf{0.0005$\pm${\scriptsize0.0003}} & \textbf{0.0005$\pm${\scriptsize0.0003}}
			& \textbf{0.0284$\pm${\scriptsize0.0114}} & \textbf{0.0304$\pm${\scriptsize0.0114}} \\
			\textbf{DDML-MLP}
			& \textbf{0.1220$\pm${\scriptsize0.1374}} & \underline{0.1785$\pm${\scriptsize0.1374}}
			& \underline{0.0007$\pm${\scriptsize0.0005}} & \underline{0.0007$\pm${\scriptsize0.0005}}
			& \underline{0.0354$\pm${\scriptsize0.0111}} & \underline{0.0370$\pm${\scriptsize0.0111}} \\
			
			\bottomrule
	\end{tabular}}\\[4pt]
	\begin{minipage}{0.89\linewidth}
		\footnotesize\raggedright
		{\footnotesize **DDML achieves average improvements of \textbf{32.2\%} in MAE and \textbf{29.3\%} in RMSE compared with the best baseline method.}
	\end{minipage}
\end{table*}

\subsection{Experiment Results}
Sections \ref{res:syn_data}-\ref{res:ps} present comprehensive experimental evaluations on synthetic, semi-synthetic, and real-world datasets, followed by visualization analysis, ablation study, and parameter sensitivity analysis.

\subsubsection{Results on Synthetic Data}\label{res:syn_data} Tables~\ref{tab:bi_data} and~\ref{tab:con_data} show the results on synthetic datasets with varying feature dimensions for the binary  and continuous treatment settings, respectively.

As shown in Table~\ref{tab:bi_data}, under the binary-treatment setting, DDML consistently outperforms traditional methods and maintains clear advantages even in high-dimensional settings. Compared with representation learning methods, DDML achieves the best performance on nearly all datasets, except that DeR-CFR obtains lower MAE on the 100-dimensional dataset. DDML also shows clear improvements over debiased estimation methods, especially in low-dimensional cases. These results indicate that the proposed disentanglement and residual orthogonalization effectively reduce the estimation bias caused by mixed covariates.

As shown in Table~\ref{tab:con_data}, for continuous treatments, we compare against baselines that natively support continuous treatments (OLS-1, BART, DML, and C-DML). Under the continuous-treatment setting, DDML achieves the best performance across all datasets, substantially outperforming both traditional methods and debiased estimation methods. Although the performance of all methods generally declines as the feature dimension increases, DDML remains consistently superior. These results demonstrate that the proposed approach is not only robust and effective under complex high-dimensional confounding, but also more broadly applicable to continuous-treatment scenarios.

\subsubsection{Results on Semi-synthetic and Real-world Data}\label{res:semi_syn_data} Table~\ref{tab:real_results} presents the performance of all methods on the semi-synthetic \textit{IHDP} and \textit{Twins} datasets, as well as the real-world \textit{Jobs} dataset. On \textit{IHDP}, DDML-MLP achieves the best MAE, while OLS-2 gives the best RMSE, indicating that DDML remains highly competitive on this dataset. On \textit{Twins}, DDML-RF ranks first on both MAE and RMSE, with DDML-MLP achieving the second-best results, showing clear superiority over all baselines. On \textit{Jobs}, DDML-RF again performs best and DDML-MLP remains among the top methods, further demonstrating the robustness and practical effectiveness of the proposed approach. These results verify that the proposed disentanglement and residual orthogonalization design generalizes well across different data settings.

\subsubsection{Visualization Analysis}\label{res:vs}
To provide more intuitive evidence for the effectiveness of DDML, we conduct a visualization analysis to verify the effects of causal-role disentanglement and residual dependence orthogonalization. We first use probing heatmaps to examine whether the learned latent subspaces demonstrate clear role specialization, validating the CRD module. We then analyze the dependence between the treatment residual and the remaining outcome error to assess whether the RDO module improves orthogonality in practice. These analyses are conducted on the \textit{Binary\_100}, \textit{Continuous\_100}, and \textit{IHDP} datasets.

\begin{figure}[ht]
	\centering
	\subfloat[\rmfamily]{\includegraphics[width = 0.16\textwidth]{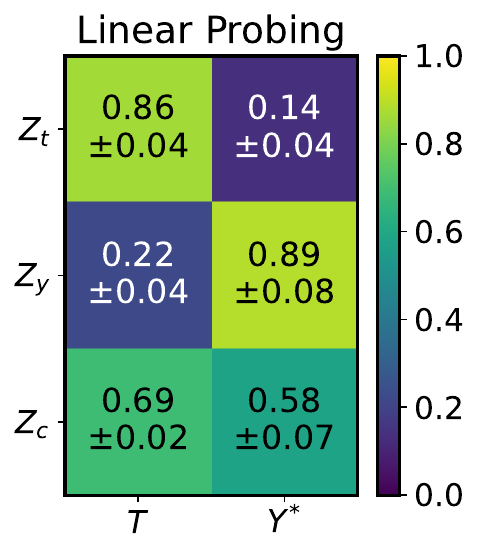}}
	\subfloat[\rmfamily]{\includegraphics[width = 0.16\textwidth]{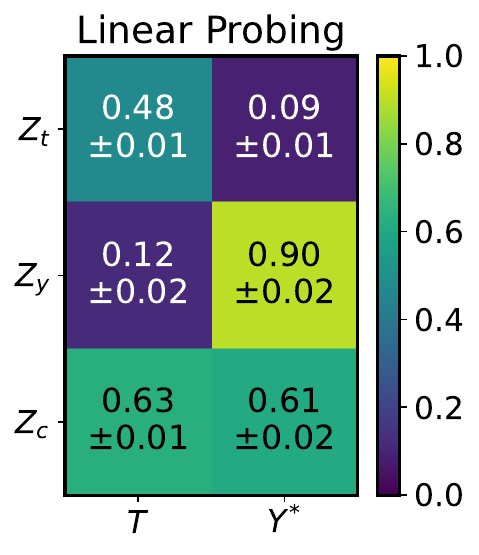}}
	\subfloat[\rmfamily]{\includegraphics[width = 0.16\textwidth]{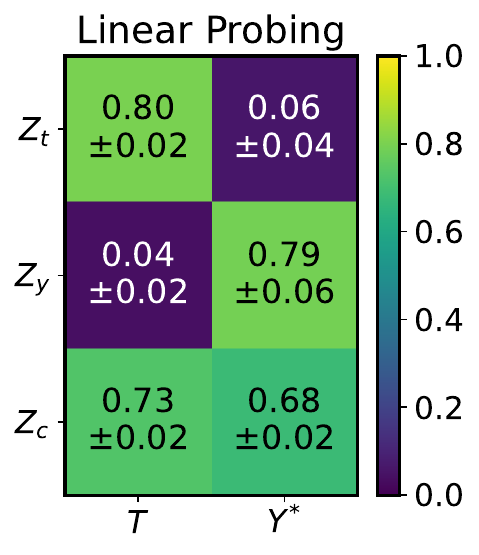}}
	\captionsetup{
		singlelinecheck=false,
		justification=justified,
		format=plain
	}
	\caption{Linear probing results on datasets of (a) \textit{Binary\_100}, (b) \textit{Continuous\_100}, and (c) \textit{IHDP}. Each row corresponds to a latent subspace ($Z_t$, $Z_y$, $Z_c$), and each column reports predictability of treatment $T$ and residualized outcome $Y^{*}$.}
	\label{fig:heatmap}
\end{figure}

(i) We use the encoder trained on the training fold to extract $(Z_t,Z_y,Z_c)$ for the corresponding held-out samples, and then fit simple linear probes on each subspace to predict the treatment $T$ and the residualized outcome $Y^{*}$. Here, $Y^{*} = Y  - \hat \theta(T - m(Z))$ removes the direct contribution of the treatment residual from $Y$, so that probing on $Y^{*}$ better reflects outcome-specific information encoded in each subspace. Treatment-related information is quantified by AUC on \textit{Binary\_100} and \textit{IHDP}, and by Spearman's rank correlation on \textit{Continuous\_100}, while outcome-related information is measured by the $R^2$ of regressing $Y^{*}$ on each subspace. As shown in Fig.~\ref{fig:heatmap}, the learned subspaces show a consistent specialization pattern across datasets. $Z_t$ is more predictive of $T$ than $Y^{*}$, $Z_y$ shows the reverse, and  $Z_c$ remains moderately predictive of both, confirming that the CRD module learns causal role-guided latent subspaces that effectively separate covariates into treatment-specific, outcome-specific, and confounding information while reducing information leakage.

\begin{figure}[ht]
	\centering
	\subfloat[]{
		\includegraphics[width=0.94\linewidth]{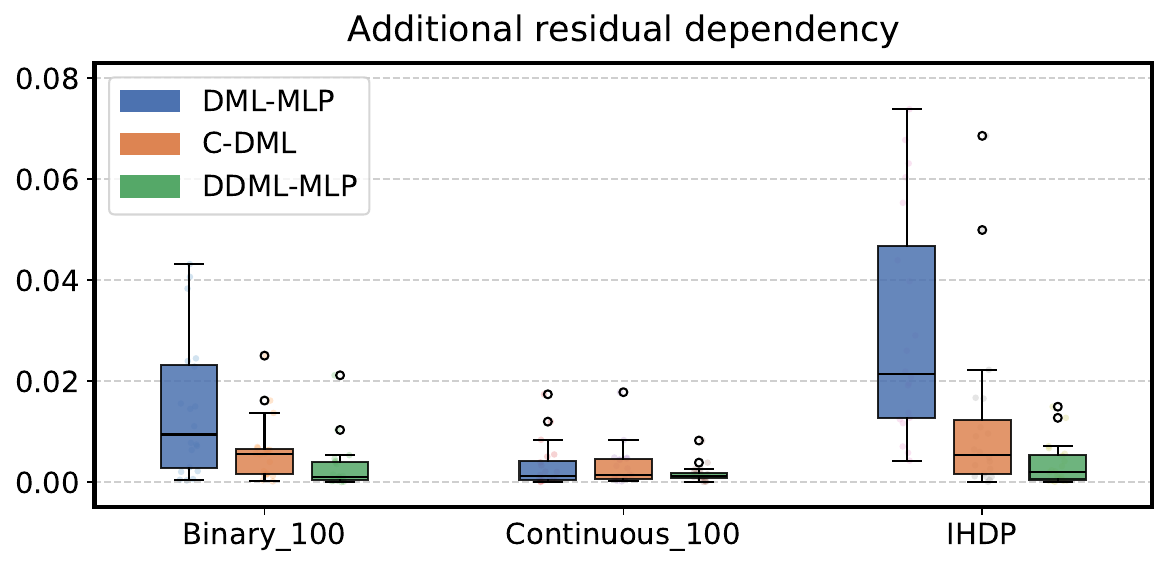}
		\label{fig:Res_corr_MLP}
	}

	\subfloat[]{
		\includegraphics[width=0.94\linewidth]{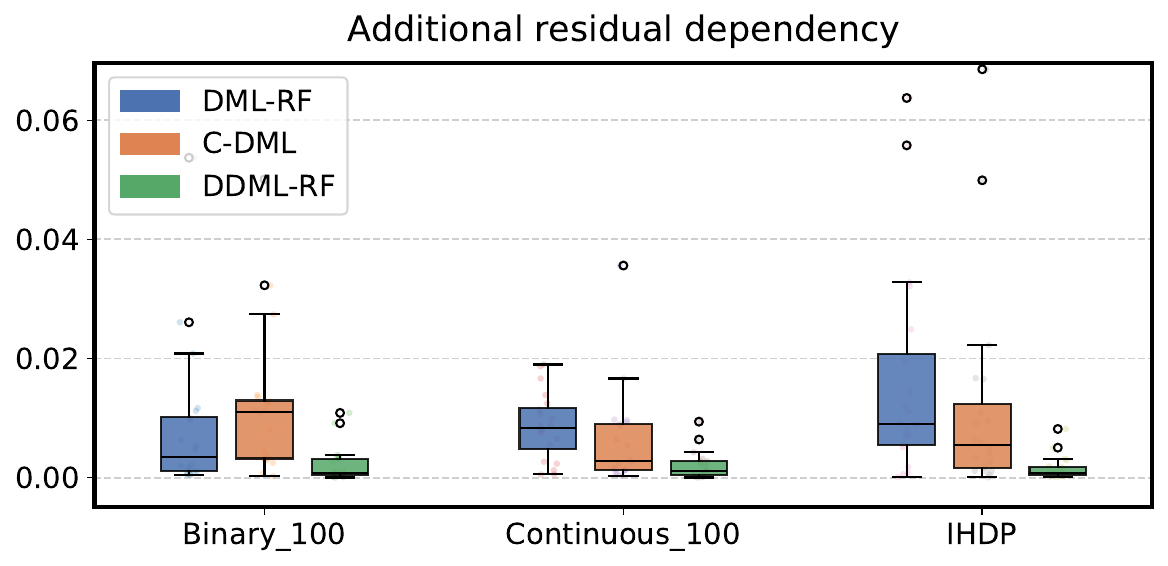}
		\label{fig:Res_corr_RF}
	}
	\captionsetup{singlelinecheck=false, justification=justified}
	\caption{Additional dependency between the treatment residual and the remaining outcome error after removing the treatment effect across \textit{Binary\_100}, \textit{Continuous\_100}, and \textit{IHDP} datasets of DDML with MLP and RF nuisance estimators.}
	\label{fig:Res_corr_all}
\end{figure}

(ii) We construct the treatment residual $\tilde{T}=T-\hat{m}(Z)$ and the outcome error $\epsilon=\tilde{Y}-\hat{\theta} \tilde{T}$, and measure their absolute Pearson correlation. As shown in  Fig.~\ref{fig:Res_corr_all}, DDML-MLP and DDML-RF consistently achieve the lowest dependence between $\tilde{T}$ and $\epsilon$ across all datasets, with smaller medians, narrower spreads, and fewer extreme values than DML and C-DML. The improvement is especially pronounced on \textit{IHDP}, where DDML markedly reduces both the magnitude and variability of residual correlation.
Notably, although C-DML often yields lower residual correlation than DML, this does not result in more accurate causal effect estimation. This is because C-DML focuses more on coordinated optimization of the residual dependence term, which can directly reduce the correlation metric itself. In contrast, DDML improves both nuisance estimation and residual orthogonality, leading to more accurate causal effect estimation.

\begin{figure}[ht]
	\centering	
	\subfloat[\rmfamily]{\includegraphics[width=0.49\linewidth]{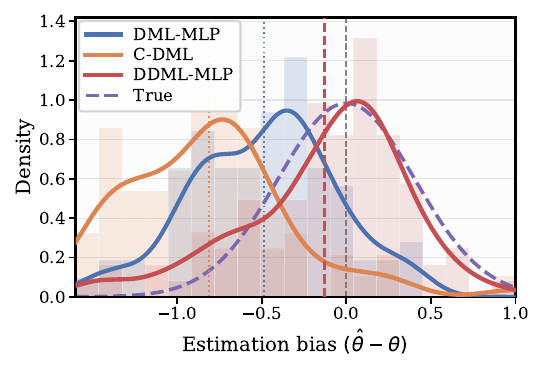}}
	\subfloat[\rmfamily]{\includegraphics[width=0.49\linewidth]{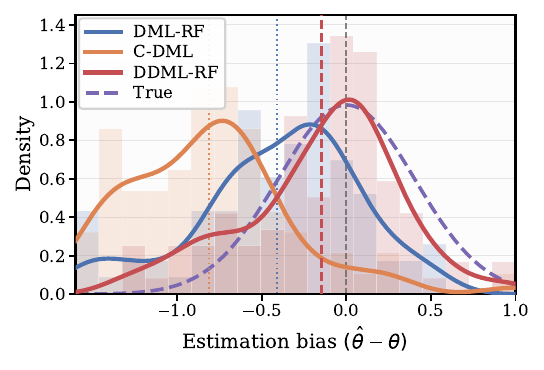}}
	\captionsetup{singlelinecheck=false, justification=justified}
	\caption{Distribution of causal effect estimation bias for DML, C-DML, and DDML on the \textit{IHDP} dataset with MLP and RF nuisance estimators.}
	\label{fig:bias_MLP_RF}
\end{figure}

The above results show that the representation learned by DDML can effectively separate causal role information, while the residual correlation analysis further demonstrates that such disentanglement significantly weakens the dependence between the treatment residual and the remaining outcome error. As shown in Fig. \ref{fig:bias_MLP_RF}, on the \textit{IHDP} dataset, the causal effect estimates produced by DDML are closer to the true causal effect and exhibit smaller bias and lower variability than those of DML and C-DML. Consequently, DDML can effectively mitigate the estimation bias caused by mixed covariates and insufficient residual orthogonality.

\subsubsection{Ablation Study}\label{res:as} 
\begin{figure}[htbp]
	\centering
	\subfloat[]{
		\includegraphics[width=0.94\linewidth]{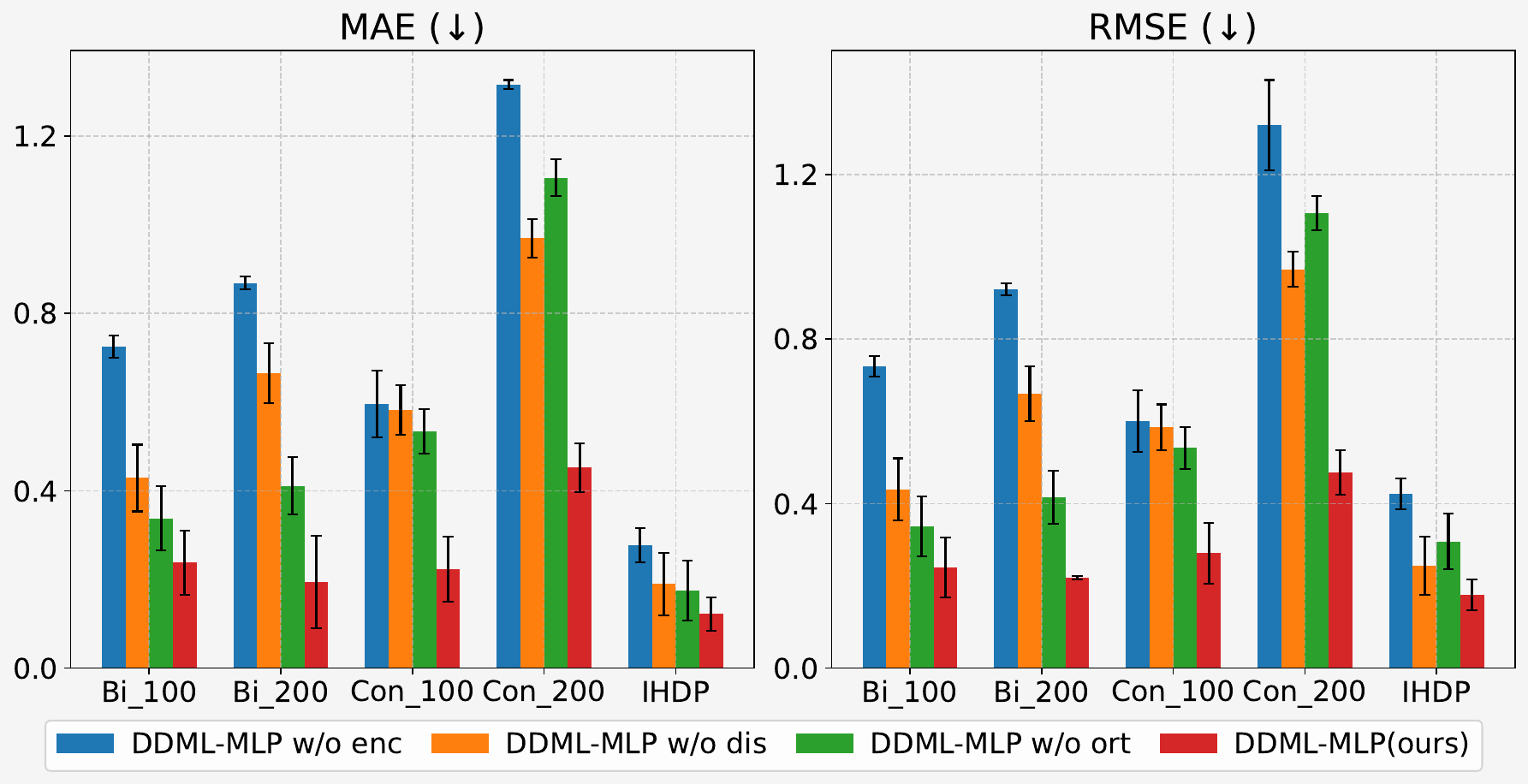}
	}
	
	\subfloat[]{
		\includegraphics[width=0.94\linewidth]{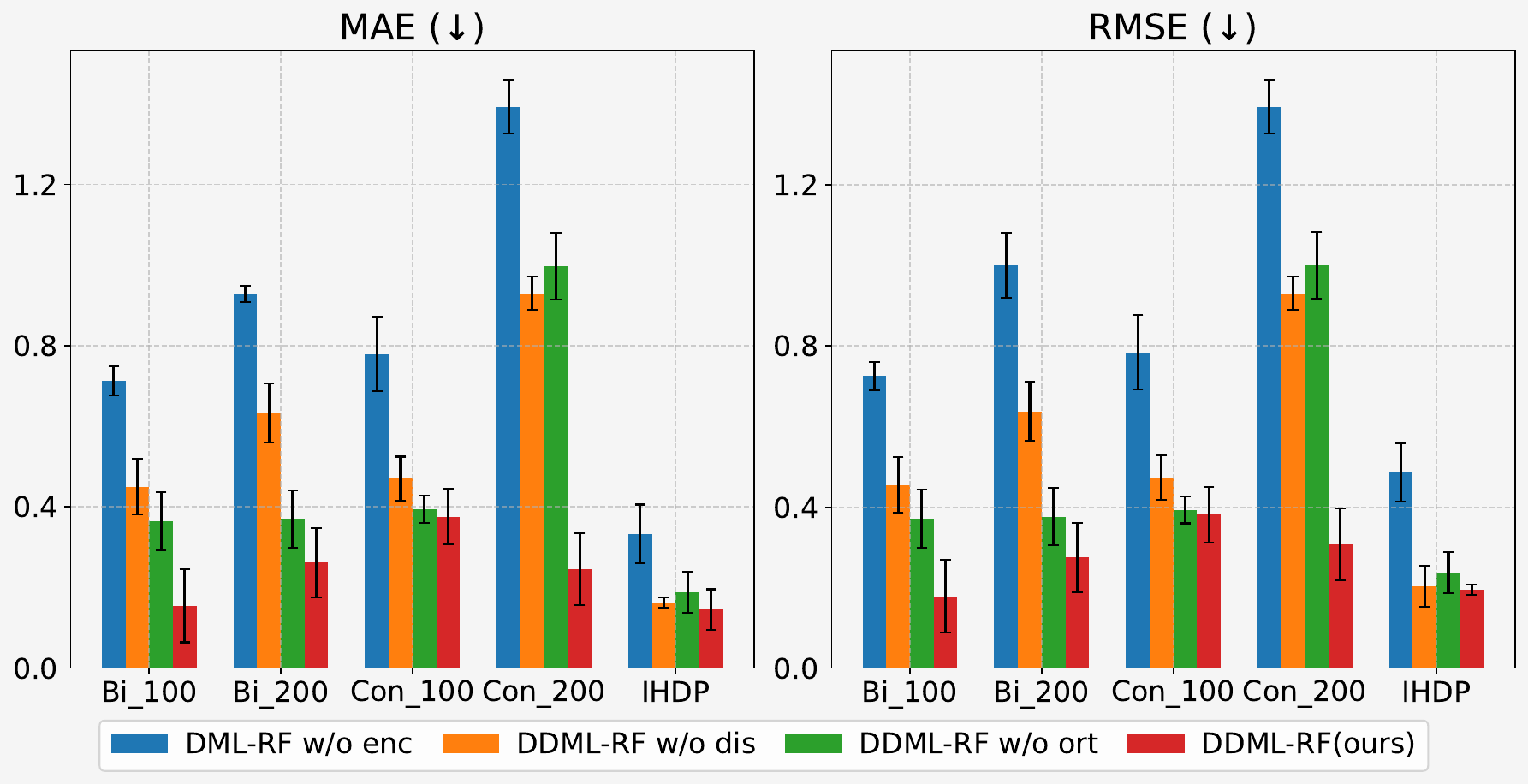}
	}
	
	\captionsetup{singlelinecheck=false, justification=justified}
	\caption{Ablation study on \textit{Binary\_100}, \textit{Binary\_200}, \textit{Continuous\_100}, \textit{Continuous\_200}, and \textit{IHDP} datasets of DDML with MLP and RF nuisance estimators.}
	\label{fig:as_all}
\end{figure}

To evaluate the contribution of each module in DDML, we conduct systematic ablation studies. Specifically, we consider three variants: \textit{DDML w/o enc} (removing the encoder), \textit{DDML w/o dis} (removing the disentanglement loss), and \textit{DDML w/o ort} (removing the orthogonalization loss). We conducted experiments on \textit{Binary\_100}, \textit{Binary\_200}, \textit{Continuous\_100}, \textit{Continuous\_200}, as well as the \textit{IHDP} dataset. The results of DDML are shown in Fig. \ref{fig:as_all}.

As shown in Fig. \ref{fig:as_all}, the complete DDML-MLP and DDML-RF consistently achieves the lowest MAE and RMSE across all datasets. Removing any module leads to a significant performance degradation, confirming that each module contributes to the final effect estimation. Among the ablations, removing the disentanglement loss usually causes the largest drop, especially in high-dimensional settings, indicating that causal role disentanglement is essential for accurate nuisance estimation. Removing the orthogonalization constraint or the encoder also degrades performance, showing that residual dependence control and representation learning are both necessary for stable estimation. Overall, the ablation results verify that the three components are complementary and jointly responsible for the robustness of DDML.

\begin{table*}[htbp]
	\centering
	\footnotesize
	\setlength{\tabcolsep}{4.8pt}
	\renewcommand{\arraystretch}{1.2}
	\caption{Cross-domain accuracy on \textit{Amazon Reviews} dataset. (Best result is highlighted in bold.)}
	\begin{tabular}{l|cccccccccccccc}
		\toprule
		& B$\to$D & B$\to$E & B$\to$K & D$\to$B & D$\to$E & D$\to$K & E$\to$B & E$\to$D & E$\to$K & K$\to$B & K$\to$D & K$\to$E & Avg & \underline{Avg. Rank} \\
		\midrule
		DGBR     & 0.6693 & 0.6582 & 0.6473 & 0.5975 & 0.5856 & 0.6108 & 0.5875 & 0.6253 & 0.6908 & 0.5775 & 0.5918 & 0.6752 & 0.6264 & 7.00 \\
		EAMB     & 0.7469 & 0.7252 & 0.7539 & 0.7250 & 0.7472 & 0.7492 &0.6810 & 0.6968 & 0.8090 & 0.6850 & 0.7190 & 0.8028 & 0.7368 & 3.71 \\
		CVS      & 0.7569 & 0.7307 & 0.7704 & 0.7300 & 0.7007 & 0.7429 & 0.6355 & 0.6378 & 0.7554 & 0.6850 & 0.6983 & 0.8023 & 0.7205 & 4.83 \\
		PCFS     & 0.7539 & 0.6872 & 0.7269 & 0.7160 & 0.7202 & 0.7434 & 0.6465 & 0.6918 & 0.7969 & 0.6840 & 0.7294 & 0.7963 & 0.7244 & 4.83 \\
		CIFD     & 0.7674 &0.7382 & 0.7709 &0.7300 & 0.7297 & 0.7484 & 0.6400 &0.7084 & 0.8024 & 0.6755 & 0.7264 & 0.7923 & 0.7358 & 3.79 \\
		\midrule
		DDML-RF  & \textbf{0.7824} & 0.7352 & 0.7729 & \textbf{0.7460} & \textbf{0.7528} & \textbf{0.7739} & \textbf{0.6855} & 0.7064 & \textbf{0.8294} & 0.6805 & 0.7284& \textbf{0.8088} & \textbf{0.7502} & \underline{\textbf{1.92}} \\
		DDML-MLP & 0.7739 & \textbf{0.7417} & \textbf{0.7854} & 0.7340 & 0.7247 &0.7699 & 0.6715 & \textbf{0.7214} & 0.8044 & \textbf{0.6885} & \textbf{0.7339} &0.8083 & 0.7462 & \underline{\textbf{1.92}} \\
		\bottomrule
	\end{tabular}\\[4pt]
	\begin{minipage}{0.975\linewidth}
		\footnotesize\raggedright
		{\footnotesize \textbf{**Note:} DDML-RF/MLP denote DDML-based top-$K$ effect-ranked causal feature selection using RF/MLP nuisance learners.}
	\end{minipage}
	\label{tab:amazon}
\end{table*}
\subsubsection{Parameter Sensitivity Analysis}\label{res:ps}
\begin{figure}[h]
	\centering
	\subfloat[]{
		\includegraphics[width=1\linewidth]{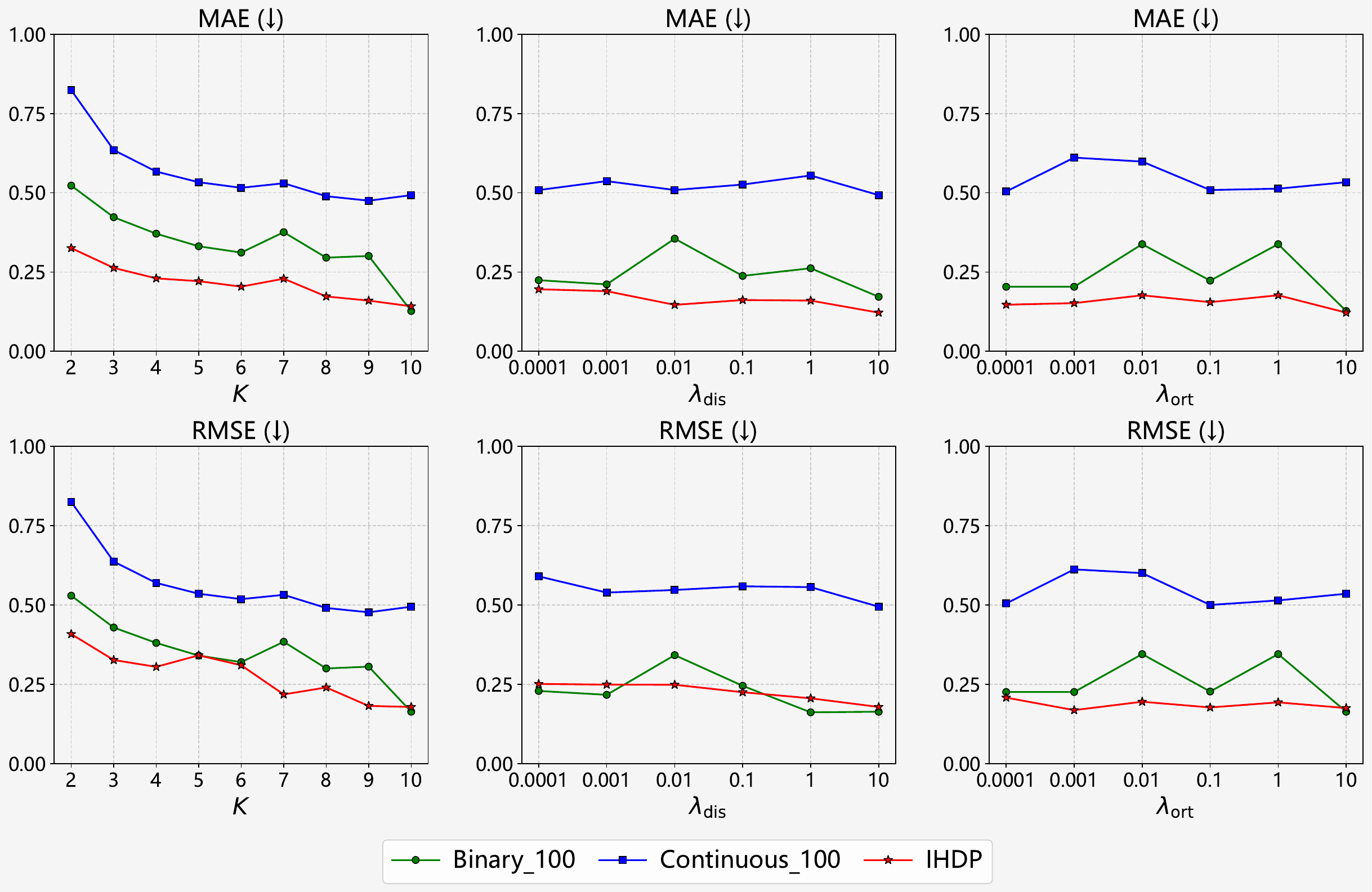}
		\label{fig:ps_mlp}
	}
	
	\subfloat[]{
		\includegraphics[width=1\linewidth]{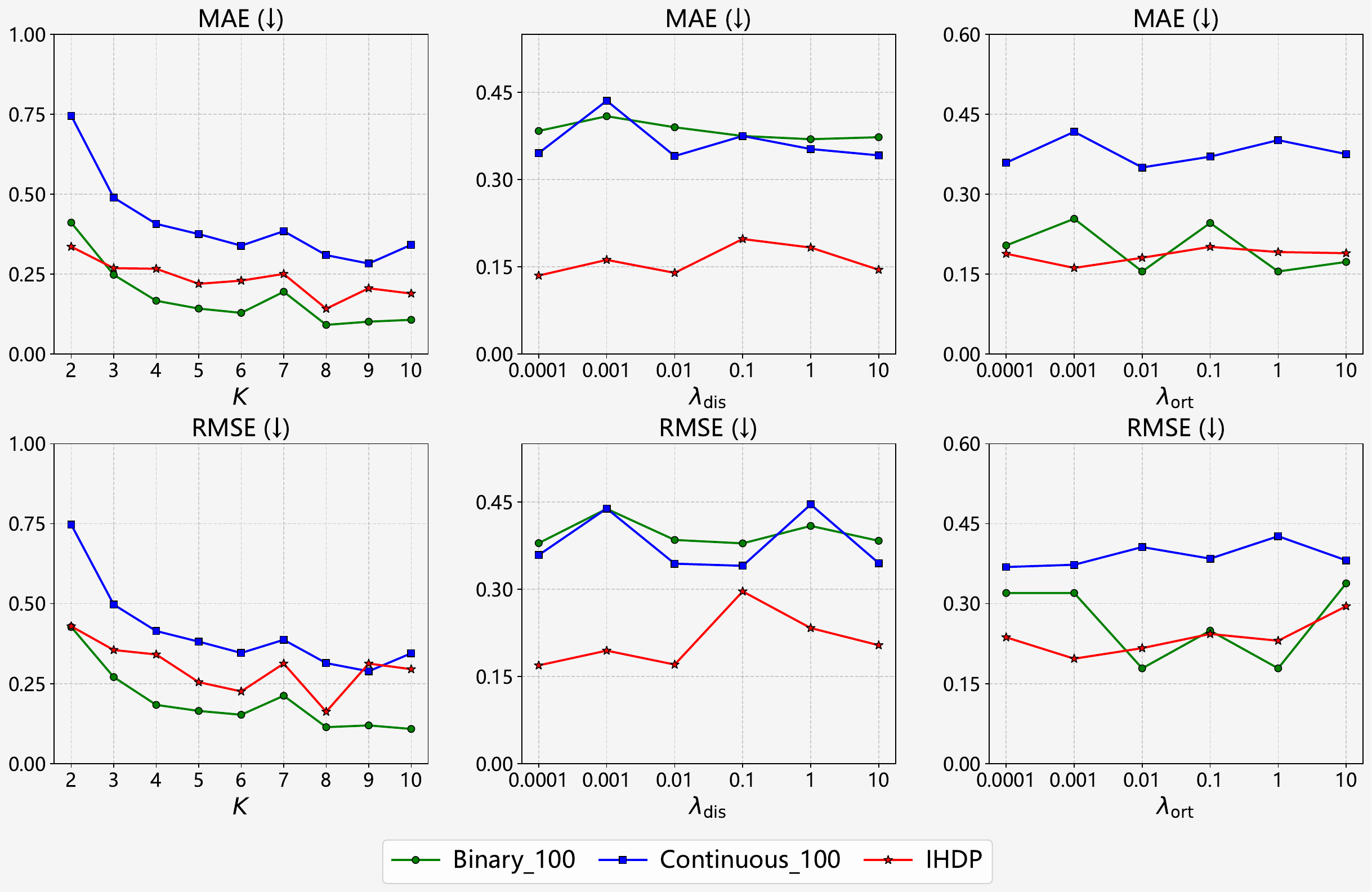}
		\label{fig:ps_rf}
	}
	
	\captionsetup{singlelinecheck=false, justification=justified}
	\caption{Sensitivity analysis of MAE and RMSE with respect to $K$, $\lambda_{\mathrm{dis}}$, and $\lambda_{\mathrm{ort}}$ on \textit{Binary\_100}, \textit{Continuous\_100}, and \textit{IHDP} of DDML. (a) MLP for nuisance estimation. (b) RF for nuisance estimation}
	\label{fig:ps_all}
\end{figure}
We perform a parameter sensitivity analysis to evaluate the influence of key hyperparameters on model performance, focusing on the number of cross-fitting folds $K$ and the two loss weights $\lambda_{\mathrm{dis}}$ and $\lambda_{\mathrm{ort}}$. We conduct experiments on \textit{Binary\_100}, \textit{Continuous\_100}, and \textit{IHDP} datasets. The results for DDML are shown in Fig.~\ref{fig:ps_all}.

(i) To study the effect of the number of cross-fitting folds ($K$), we vary $K$ from 2 to 10 while fixing all other hyperparameters. As shown in Fig.~\ref{fig:ps_all}, both MAE and RMSE generally decrease as $K$ increases from small to moderate values, indicating that additional folds help alleviate nuisance estimation bias. As $K$ further increases, both metrics tend to stabilize with only minor fluctuations. These results suggest that moderate values of $K$ are sufficient to achieve stable and accurate estimation across both settings and datasets.

(ii) To study the effect of the two loss weights ($\lambda_{\mathrm{dis}}$ and $\lambda_{\mathrm{ort}}$), we vary both loss weights from 0.0001 to 10 while keeping all other hyperparameters fixed. As shown in Fig.~\ref{fig:ps_all}, MAE and RMSE remain largely stable, with noticeable degradation only at extreme values. These results indicate that our method is not highly sensitive to these loss weights, and stable performance can be achieved without extensive tuning.

\section{Real Data Application}\label{R_A}
To further validate the effectiveness of DDML in estimating causal effects, we extend DDML to an out-of-distribution (OOD) generalization task through causal feature selection. Specifically, we treat each feature as a candidate treatment variable and use DDML to estimate its effect on the target label while controlling for the remaining covariates. We then select features with the largest estimated effects that are also stable across source domains, and train a downstream classifier using only the selected features. This setting evaluates whether DDML can identify causal features that remain stable across domains and improve OOD generalization performance.

\subsection{Experimental Setup}
We evaluate our method on the real-world \textit{Amazon Reviews} dataset with four domains (Books, DVDs, Kitchen appliances, and Electronics). Each domain contains approximately 2,000 samples, 400 features, and one label variable. We adopt preprocessed binary sentiment datasets and treat each domain as a distinct environment \cite{wang2018visual}. Following \cite{yangCausal2023}, we adopt a single-source identification and multi-target validation protocol: we select one domain as the source, estimate feature-level causal effects and select causal features on the source, then directly evaluate on the other three domains, yielding $4\times3=12$ cross-domain transfer tasks. DDML is used to estimate the causal effect of each feature; features are ranked according to the magnitude of their estimated effects, and the top-$K$ ranked features are used to train downstream classifiers. We conduct comprehensive experiments and compare DDML with five state-of-the-art causal feature selection methods. DGBR \cite{kuang2018stable} leverages sample reweighting to eliminate variable correlations for stable prediction across unknown environments. EAMB \cite{guo2022error} adopts an error-aware Markov blanket learning algorithm to enhance the robustness of causal feature discovery. CVS \cite{kuang2023stable} utilizes seed variables to guide the learning of invariant causal structures for robust generalization. PCFS \cite{yangCausal2023} employs a progressive learning strategy to address the challenge of causal feature selection under sample selection bias. CIFD \cite{wangDiscovering2024} focuses on identifying causally invariant features to improve OOD generalization performance. We use classification accuracy as the evaluation metric, reporting the result for each transfer task as well as the average across all tasks.

\subsection{Experimental Results}
As shown in Table \ref{tab:amazon}, DDML-RF achieves the best average accuracy (0.7502) and the best average rank (1.92), while DDML-MLP attains a comparable average rank (1.92) with a slightly lower average accuracy (0.7462). Across the 12 transfer tasks, methods based on DDML-RF and DDML-MLP achieve the best or second-best accuracy in most cases, demonstrating the effectiveness of causal feature selection for OOD generalization. In contrast, existing baselines (DGBR, EAMB, CVS, PCFS, CIFD) are clearly weaker, with average accuracies ranging from 0.6264 to 0.7368 and substantially worse average ranks, and DGBR performing the worst (Avg = 0.6264, Avg Rank = 7.00). These results indicate that ranking and selecting features by causal effects estimated by DDML can identify more stable features, leading to more reliable cross-domain prediction.

\subsection{Statistical Test}\label{Statistical_Test}
To comprehensively evaluate the performance of our algorithm on the real-world dataset, we adopt non-parametric statistical tests to compare DDML-based methods with competing methods across all tasks. We first apply the Friedman test at the 0.05 significance level, with the null hypothesis that all methods have identical average ranks. We then use the Nemenyi test \cite{demsar2006statistical} for pairwise comparisons, declaring two methods significantly different if their average-rank difference exceeds the critical difference (\textit{CD}). The \textit{CD} for the Nemenyi test is computed as:
\begin{equation}
	\mathit{CD}=q_{\alpha,\Delta}\sqrt{\frac{\Delta(\Delta+1)}{6\eta}},
\end{equation}
where $\Delta$ is the number of compared methods, $\eta$ is the number of tasks, $\alpha$ is the significance level, and $q_{\alpha,\Delta}$ is the critical value from the Studentized range statistic. In our setting, we compare $\Delta=7$ methods on $\eta=12$ transfer tasks at $\alpha=0.05$, yielding $\mathit{CD}=2.60$.
\begin{figure}[htbp]
	\centering
	\includegraphics[width=0.9\linewidth]{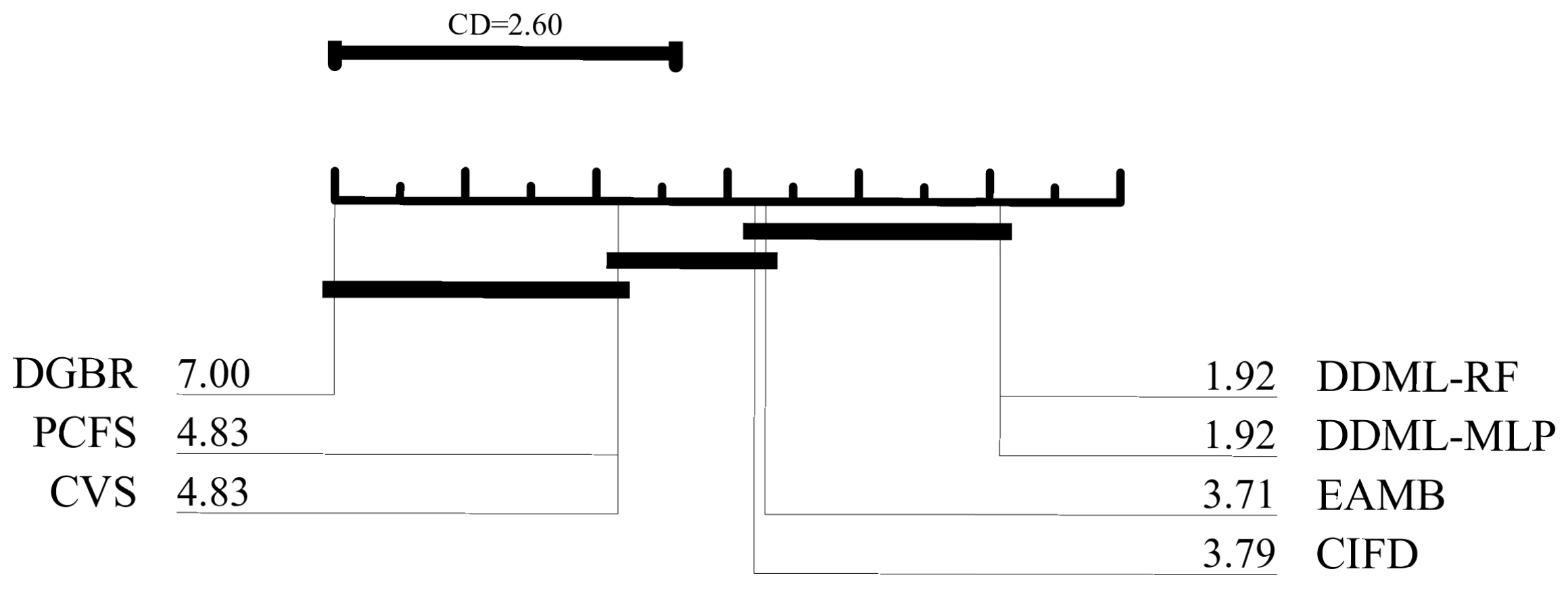}
	\caption{\textit{CD} diagram of the Nemenyi test for the Accuracy metric
		on the \textit{Amazon Reviews} dataset (the lower the rank value, the better the performance). }
	\label{fig:st}
\end{figure}

Fig.~\ref{fig:st} shows the \textit{CD} diagram of the Nemenyi test based on average ranks over the 12 cross-domain transfer tasks (lower is better). DDML-RF and DDML-MLP achieve the best ranks (both 1.92), forming the leading group in the diagram. Their rank differences to the strongest baselines, such as EAMB (3.71) and CIFD (3.79), are smaller than the critical difference (CD = 2.60), indicating that these methods are statistically indistinguishable at the chosen significance level. In contrast, DGBR (7.00) is well separated from the leading group with a rank gap exceeding the CD, and is therefore significantly worse under the Nemenyi criterion. Overall, the CD analysis supports that DDML consistently ranks among the top performers across domain shifts and provides stable cross-domain generalization performance.

\section{Conclusion and Future Work}\label{C_F}
In this paper, we addressed causal effect estimation with mixed covariate roles, where DML often suffers from biased and unstable estimates. This issue mainly arises from the mixing of distinct latent factors in the covariates, which hinders reliable nuisance estimation, and from the residual dependence induced by nuisance estimation errors, which further compromises causal effect estimation. To address these challenges, we propose DDML, a novel method for accurate causal effect estimation. DDML first improves nuisance estimation by disentangling covariates into confounders, treatment-specific factors, and outcome-specific factors based on their causal roles. It further strengthens causal effect estimation by mitigating the residual dependence caused by nuisance estimation errors, improving estimation accuracy in practice. Extensive experiments on synthetic, semi-synthetic, and real-world datasets demonstrate that DDML consistently outperforms existing baselines. In future work, we will extend DDML to federated causal effect estimation, enabling privacy-preserving and communication-efficient estimation under heterogeneous, decentralized data.
 
\bibliographystyle{IEEEtran}
\bibliography{IEEEabrv,ref}

\end{document}